\documentclass[10pt]{IEEEtran}
\usepackage{amsfonts}
\usepackage{amsmath}
\usepackage{amssymb}
\usepackage{bm}
\usepackage{cite}
\usepackage{graphicx}
\usepackage{epstopdf}
\usepackage{epsfig}
\usepackage{url}
\usepackage{setspace}
\usepackage{subfigure}
\usepackage{color}
\usepackage{algorithmicx}
\usepackage{algpseudocode}
\usepackage{amsmath}
\usepackage{algorithm,times}
\usepackage{threeparttable}
\usepackage{lineno}
\usepackage{multirow}
\usepackage{verbatim}
\usepackage{booktabs}
\usepackage{textcomp}
\usepackage{siunitx}

\graphicspath{{./figures/}}

\begin{document}
\begin{spacing}{1}
\title{Mapping the Buried Cable by Ground Penetrating Radar and Gaussian-Process Regression\thanks{%
The authors are with USTC-Birmingham Joint Research Institute in
Intelligent Computation and Its Applications, School of Computer Science
and Technology, University of Science and Technology of China, Hefei
230027, China (email: zhou0612@ustc.edu.cn, qqchern@ustc.edu.cn, saintfe@mail.ustc.edu.cn, hchen@ustc.edu.cn. Corresponding author: Huanhuan Chen).
}
\author{Xiren Zhou, Qiuju Chen, Shengfei Lyu, Huanhuan~Chen,~\IEEEmembership{Senior Member,~IEEE}}
}
\maketitle
\begin{abstract}

With the rapid expansion of urban areas and the increasingly use of electricity, the need for locating buried cables is becoming urgent.
In this paper, a noval method to locate underground cables based on Ground Penetrating Radar (GPR) and Gaussian-process regression is proposed. Firstly, the coordinate system of the detected area is conducted, and the input and output of locating buried cables are determined.
The GPR is moved along the established parallel detection lines, and the hyperbolic signatures generated by buried cables are identified and fitted, thus the positions and depths of some points on the cable could be derived.
On the basis of the established coordinate system and the derived points on the cable, the clustering method and cable fitting algorithm based on Gaussian-process regression are proposed to find the most likely locations of the underground cables. Furthermore, the confidence intervals of the cable's locations are also obtained. Both the position and depth noises are taken into account in our method, ensuring the robustness and feasibility in different environments and equipments. Experiments on real-world datasets are conducted, and the obtained results demonstrate the effectiveness of the proposed method.

\end{abstract}


\section{Introduction}

Locating cables precisely has always been a prerequisite for maintaining the normal operation of the urban power system. Many underground cables are reaching the end of their practical life and need to be repaired or replaced\cite{jaw2013locational}. Existing pipeline maps could generally provide rough locations of the buried cables, but it's challenged to be specific to each point on these cables\cite{yatim2014automated}. Therefore, it is vital to locate buried cables before excavation and construction. In real-world applications, locating a buried cable could be abstracted into locating a underground curved segment, which could be divided into two parts: determine the position and depth of some points on the buried cable, and then infer the location of the whole cable based on these points.

To extract information of buried ultities in the shallow subsurface, Ground Penetrating Radar (GPR) has been widely used due to its non-destructive property \cite{daniels2004ground}.
If the cable or pipeline is buried within the effective detection depth of the GPR and has a different dielectric constant from the surrounding medium, a hyperbolic signature would be formed on the obtained GPR B-scan image after moving the GPR across the cable or pipeline\cite{butnor2001use}.
By identifying and fitting the hyperbolic signature on the B-scan image, the position and depth of the buried cable or pipeline on the cross section in this image could be estimated\cite{shihab2005radius}. There are many published methods to process and interpret hyperbolic signatures on B-scan images, including Hough-transform based methods \cite{illingworth1988survey,capineri1998advanced}, least-square methods \cite{bookstein1979fitting,akima1978method,porrill1990fitting,jaw2011accuracy}, machine-learning based methods \cite{al2000automatic,caorsi2005electromagnetic,youn2002automatic,pasolli2009automatic,gamba2000neural,delbo2000fuzzy,maas2013using}, and some combinations of the above methods that could obtain more precise results \cite{borgioli2008detection,chen2010probabilistic,chen2010robust,windsor2014data,dou2016real}. 
In \cite{chen2010probabilistic}, Chen \emph{et al.} proposed a probabilistic hyperbola mixture model to extract and fit hyperbolic point clusters on the B-scan image. The Expectation-Maximization (EM) algorithm is upgraded to extract points on multiple hyperbolic signatures from a GPR image, which is then fitted to estimate the equation of each hyperbolic point set. In \cite{windsor2014data}, the position and time data are pared and recorded in a generalized pair-labeled Hough transform, and a conventional least-square method is then utilized to infer position, depth, and radius of the buried pipeline. In\cite{dou2016real}, the Column-Connection Clustering (C3) algorithm is proposed to scan the B-scan image and separate point clusters with intersections. Point clusters with hyperbolic signatures are then identified by a neural-network-based method. In our previous work\cite{zhou2018automatic}, a GPR B-scan image interpreting model has been proposed. The model could estimate the radius and depth of the buried pipelines by converting the GPR B-scan images into binary ones, scanning the binary image to cluster points, and fitting the obtained point clusters with hyperbolic signatures. Experiments on cement and metal pipes have been conducted and the results validated the accuracy and efficiency of this model. Subsequently in \cite{zhou2019efficient}, the model is further extended to estimate insulated pipes in the soil by combining GPR with Electric-Field (EF) methods.
By applying the above methods, the depth and position of some points on the buried cable or pipeline could be roughly derived from GPR B-scan images.

Once the cable information at some points are derived, a challenging work is to infer the locations of the cables in the detected area from these individual hypothsized points. 
In \cite{dou20163d}, a Marching-Cross-Sections (MCS) algorithm is proposed to merge the individual hypothesized pipeline information from multiple sensors and locate underground pipeline segments. Detected points on the pipelines or cables are connected by a extended Kalman Filter (KF) \cite{jazwinski2007stochastic}  with straight line segments, assisted by some rules that manage buried utilities to keep potentially existing ones and to discard invalid ones. 
In \cite{chen2011buried}, the directions of buried objects are roughly determined by the existing pipeline map. Multiple GPR detections at different directions are then conducted to derive the specific location of each buried object. 
In \cite{zhou2019probabilistic}, a probabilistic mixture model is conducted to denoise and classify the data from detected points, which are then fitted by a Classification Fitting Expectation Maximization (CFEM) algorithm to locate the buried pipeline. 
The above methods are mainly aimed at mapping buried pipelines that are straight. However, in order to avoid underground obstacles, buried cables might be bent, as well as the use of pipe-jacking technology\footnote{Pipe-jacking is a trenchless technology for buried utilities.}\cite{ma2008development}, thus describing the locations of buried cables by straight line segments might lead to errors. 
To address this issue, in \cite{jiang2019cable}, a Three-Dimensional Spline Interpolation (TDSI) algorithm is proposed. Detected points on the cable obtained from GPR are interpolated with a smooth curve. The main concern of this algorithm is that the cable information obtained by GPR and positioning equipment could be inaccurate.
In practical applications, the obtained GPR data could be noisy due to the system noise, the heterogeneity of the underground medium, and mutual wave interactions \cite{daniels2004ground}, thus the interpreted information of the buried cable could be with errors. The positioning accuracy would be affected by the utilized equipment and the surrounding buildings.
Both the position and depth noises should be taken into account to obtain the more reliable locations and the most likely intervals of the buried cables.

\begin{figure}[htbp]
	\centering
	\includegraphics[width=0.49\textwidth]{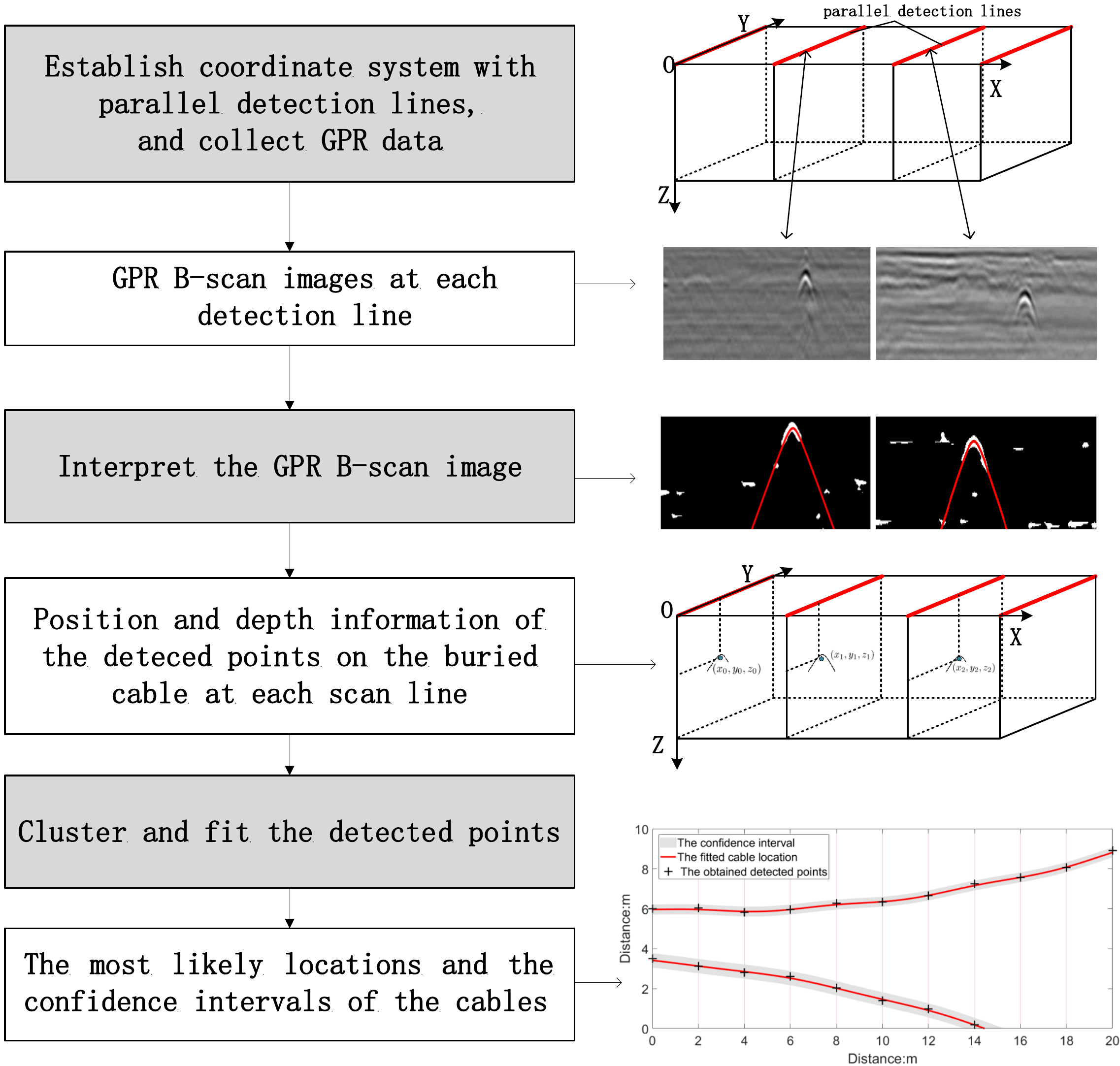}
	\caption{The procedure of the proposed method. The gray blocks refer to processing actions which lead to various status illustrated in white rectangular boxes. Right the white rectangular box is the diagram of each status.}
	\label{Procedure}
\end{figure}

In this paper, a novel method to locate buried cables is proposed, of which the procedure is presented in Fig. \ref{Procedure}.
The in/output of locating the buried cables are determined by conducting a coordinate system with parallel detection lines\footnote{A detection line indicates a line segment on a ground surface where the GPR is moved to collect the B-scan image.}, along which the GPR is moved to collect B-scan images. The obtained images are processed to extract hyperbolic point clusters by extending part of our previous work\cite{zhou2018automatic}, which are fitted to estimate the positions and depths of some points on the buried cables at each detection line. Instead of connecting the obtained points with straight line segments or interpolate them with curves, these points are clustered and processed by a designed cable fitting algorithm based on Gaussian-process regression to obtain the most likely locations of the underground cables. The confidence intervals of the buried cables are also obtained, which could provide early warning for the excavation, and could also reduce the range of precise detections. In our method, both the position and depth noises are considered, which improves the robustness in different environments. The main contributions of this paper could be summarized as follows:
\begin{enumerate}
	\item The in/output of locating buried cables are determined by conducting a coordinate system of the detected area, normalizing the data set composed of detected points, and describing the locations of the cables.
	\item A hyperbolic fitting algorithm is designed for the signature on B-scan images generated by cables, where some parameters of the standard hyperbolic equation are simplified. Furthermore, on the basis of Restricted Algebraic-Distance-based Fitting algorithm (RADF)\cite{zhou2018automatic}, the hyperbolic point clusters are fitted with more accurate equations through several iterations that minimize the sum of orthogonal distance.	
	\item A cable fitting algorithm based on Gaussian-process regression is proposed, which takes both depth and position noises of each detected point into account. The algorithm could provide the most likely locations and confidence intervals of the buried cables.
\end{enumerate}

%As Fig. \ref{Procedure1} demonstrated,
%\begin{figure*}[htbp]
%	\centering
%	\includegraphics[height=1.05in]{Procedure}
%	\caption{The procedure of the proposed algorithm. The gray blocks refer to processing actions which lead to various status of the data illustrated in white rectangular boxes. Below the white rectangular box is the diagram of each status.}
%	\label{Procedure1}
%\end{figure*}

%the Three-Dimensional Spline Interpolation (TDSI) algorithm is used to interpolate those points with a smooth curve, thus the locations and depths of the cables segment are obtained. 
 %The main contributions of this paper could be summarized as follows:
 %begin{enumerate}
 %As cables are much thinner than the pipe, Three-Dimensional Spline Interpolation (
 %\item The B-scan image with downward-opening hyperbolic signatures is identified and fitted by the GPR B-scan image processing algorithm  automatically.
 %\item TDSI algorithm was proposed to map the buried cables that takes the depth information into account compared with 2-D model. And compared with other 3-D interpolation algorithms, TDSI algorithm yields a more accurate estimation on the chosen datasets.
 %The Three-Dimensional interpolation is used to abtained a smooth 3-D curve add the depth information compared with 2-D model, which could more intuitively and accurately visualize the direction of buried cable.
%\end{enumerate}

The rest of this paper is organized as follows. The GPR B-scan image interpreting model is introduced in Section II. Section III describes the method of locating buried cables based on Gaussian-process regression, including conducting the coordinate system, clustering the detected points into different cables, and finding the most likely locations and confidence intervals of the buried cables. %The procedure of the proposed model is presented in Section IV.
Experiments are conducted and analyzed in Section IV. Finally, conclusions are drawn in Section V.

\section{The GPR B-scan image interpreting model}
\label{GPRprocess}

In this section, the theoretical model of estimating buried cables by GPR is presented, where some parameters of standard hyperbolic curves are simplified. Then the GPR B-scan image interpreting model that extends part of our previous work\cite{zhou2018automatic} is proposed. The fitting result of Restricted Algebraic-Distance-based Fitting algorithm (RADF)\cite{zhou2018automatic} is used as the initialization, and a more accurate hyperbolic equation is obtained by iteratively
minimizing the sum of orthogonal distance from points to the target hyperbola.

\subsection{The Hyperbolic Model Generated by the Buried Cable}
\begin{figure}[htbp]
	\centering
	\includegraphics[width=0.33\textwidth]{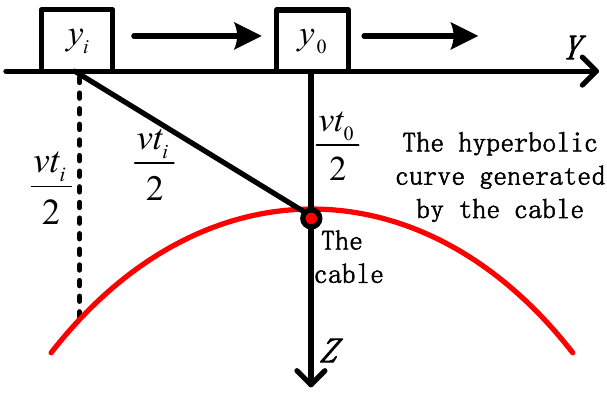}
	\caption{The hyperbolic model of buried cables. $Y$ and $Z$ are the two axises that is consistent with the coordinate system established in the next section. $y_{i}$ is the location of GPR, $t_{i}$ is the two-way travel time, and $v$ is the velocity of wave propagation. $y_{0}$ and $t_{0}$ represent the position and two-way travel time when GPR is above the pipe.}
	\label{Fit_Model}
\end{figure}
The hyperbolic signatures on the GPR B-scan image are often formulated as a geometric model, where the signal assumes the diagram of a function of the GPR position on the detected line to the two-way travel time\footnote{The two-way travel time represents the time that the wave runs from the transmitter to the object then to the receiver.} $t$ of the electromagnetic magnetic wave\cite{conyers2002ground}.

Considering the underground material to be homogeneous, $v$ is a constant value that indicates the velocity of propagation. Since cables are much thinner than pipes with non-negligible radius, the hyperbolic model conducted for pipes\cite{shihab2005radius} could be simplified by ignoring the cable's radius. As Fig. \ref{Fit_Model} shows, $y_i$ denotes the horizontal position, and $y_0$ represents the position when the GPR is above the cable. The hyperbolic signature generated by the cable could be represented as:
\begin{equation}
	\label{qc}
	(\frac{vt_i}{2})^2-(y_i-y_0)^2=	(\frac{vt_0}{2})^2,
\end{equation}
which could be converted to
\begin{equation}
	\label{hyperbolicform}
	\frac{t_i^2}{t_0^2}-\frac{(y_i-y_0)^2}{{(\frac{vt_0}{2})^2}}=1.	
\end{equation}
This is a hyperbolic equation about $t_i$ and $y_i$. By fitting the point clusters with hyperbolic equation, the position and depth of the buried cable could be derived \cite{jiang2019cable}.

\subsection{Interpreting GPR B-scan Images }

Interpreting the GPR B-scan image to estimate buried cable consists of two parts: extracting point clusters with hyperbolic signatures, and fitting the extracted point clusters to estimate the cables' position and depth.

\subsubsection{Extracting Point Clusters with Hyperbolic Signatures}
\begin{figure}[htbp]
	\centering
	\subfigure[]{ \centering
		\label{real1_ori}
		\includegraphics[height=1in]{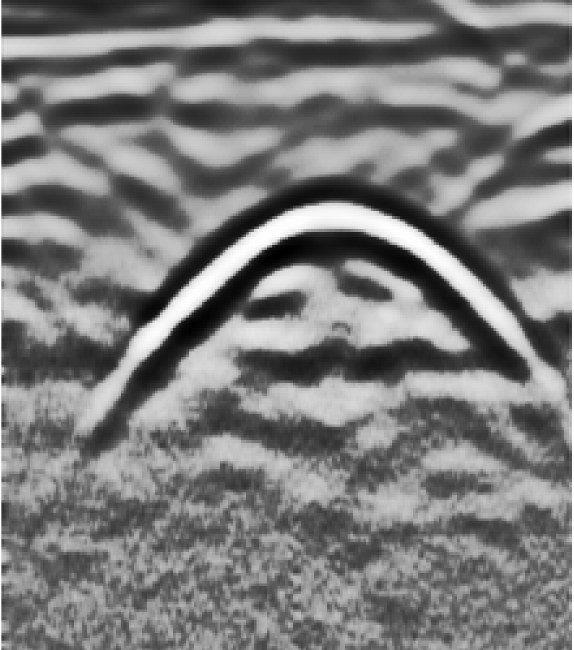}}
	\subfigure[]{ \centering
		\label{real1_oc}
		\includegraphics[height=1in]{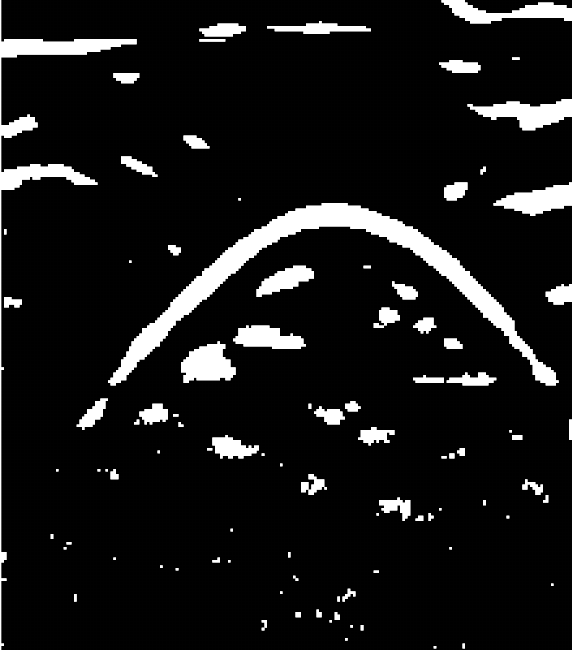}}
	\subfigure[]{ \centering
		\label{real1_OSCA}
		\includegraphics[height=1in]{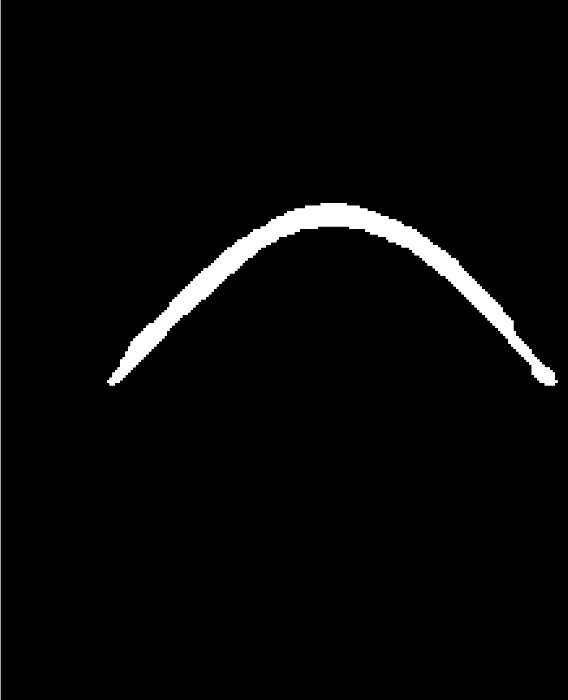}}
	\caption{These three images show the processing flow of extracting a hyperbolic point cluster from a GPR B-scan image. (a) is the original image. (b) is the preprocessed binary imaged. (c) is the obtained result of OSCA and PFJ, where the point cluster with hyperbolic signature is identified and extracted.}\label{real1_data}
\end{figure}

In our previous work\cite{zhou2018automatic}, the GPR B-scan image preprocessing method, Open-Scan Clustering Algorithm (OSCA), and Parabolic Fitting-based Judgment method (PFJ) have been proposed. The preprocessing method transforms the B-scan images into binary ones, and remove most of the discrete noises. The obtained binary image is then scanned by OSCA, and point clusters with downwardly-opening signatures are extracted. After that, PFJ is applied to further identify point clusters with hyperbolic signatures. The preprocessing method, OSCA and PFJ have been introduced in detail in \cite{zhou2018automatic}, and would not be detailed here. The process of extracting point clusters with hyperbolic signatures is illustrated in Fig. \ref{real1_data}. 
The obtained point clusters would be fitted by the algorithm introduced in the follows.

\subsubsection{Fitting the Extracted Point Clusters}
The fitting algorithm could be divided into two steps. Firstly, the RADF\cite{zhou2018automatic} is used to quickly fit the point clusters to Equation \eqref{hyperbolicform} as the initialization. After that, the initial hyperbolic equation is modified to obtain a more accurate result by minimizing the sum of orthogonal distance from points to the target hyperbola through several iterations.

The general hyperbola with focal point on the vertical axis could be presented as
\begin{equation}
	\label{generalhyp}
	\frac{(z-z_0)^2}{A^2}-\frac{(y-y_0)^2}{B^2}=1.
\end{equation}
Relating Equation \eqref{generalhyp} and \eqref{hyperbolicform}, it could be seen that $z_0=0$ in the hyperbola generated by the cable. Thus the  parametric form the hyperbola could be presented as
\begin{equation}
	\left\{  
	\begin{aligned} 
		&y=A\sinh\varphi+C,\\
		&z=B\cosh\varphi,
	\end{aligned}  
	\right.  
\end{equation}
where $C=y_0$.
Given a point cluster $(y_i,z_i)_{i=1}^m$, the orthogonal distance $d_i$ from a point $p_i=(y_i,z_i)$ to the hyperbola could be expressed by
\begin{equation}
d_i^2=\min[(y_i-y(\varphi_i))^2+(z_i-z(\varphi_i))^2],
\end{equation}
where the point $(y(\varphi_i),z(\varphi_i))$ is the nearest corresponding point of $p_i$ on the hyperbola. The $A$, $B$ and $C$ could be determined by  
\begin{equation}
	\min	\sum_{i=1}^m d_i^2,
\end{equation}
which is equivalent to solving the nonlinear least squares problem
\begin{equation}
	\label{leastsquare}
	\begin{pmatrix} y_i \\ z_i \end{pmatrix}-\begin{pmatrix} C \\ 0 \end{pmatrix}-\begin{pmatrix} A\sinh\varphi_i \\ B\cosh\varphi_i \end{pmatrix} \approx 0, \text{for }i=1,\cdots, m.
\end{equation}
Thus, there are $2m$ nonlinear equations for $m+3$ unknowns: $\varphi_1,\varphi_2,\cdots,\varphi_m,A,B,C$. To solve the minimization problem, the Gauss-Newton iteration is employed. As aforementioned, the initialization of the Gauss-Newton iteration is determined by RADF. The experimental studies in Section \ref{experiments} demonstrate that this initialization is appropriate and robust against noise. Moreover, in the conducted experiments, the accuracy of the proposed fitting algorithm is also better than RADF when fitting hyperbolic signatures generated by buried cables.

\section{Locating buried cables based on Gaussian-process regression}

In this section, the coordinate system is established with parallel detection lines. The input and output of locating buried cables are also normalized. Then the clustering method and the cable fitting algorithm are applied to estimate the location and confidence intervals of the buried cables in the detected area, provided the depths and positions of detected points at each detection line. 
%The pseudo code of the above whole process is given at the end of this section. It should be noted that the $x$, $y$ and $z$ in this section has nothing to do with the ones in the previous section.

\subsection{The Coordinate System and the In/Output of Locating Buried cables}

\begin{figure}[htbp]
	\centering
	\includegraphics[width=0.35\textwidth]{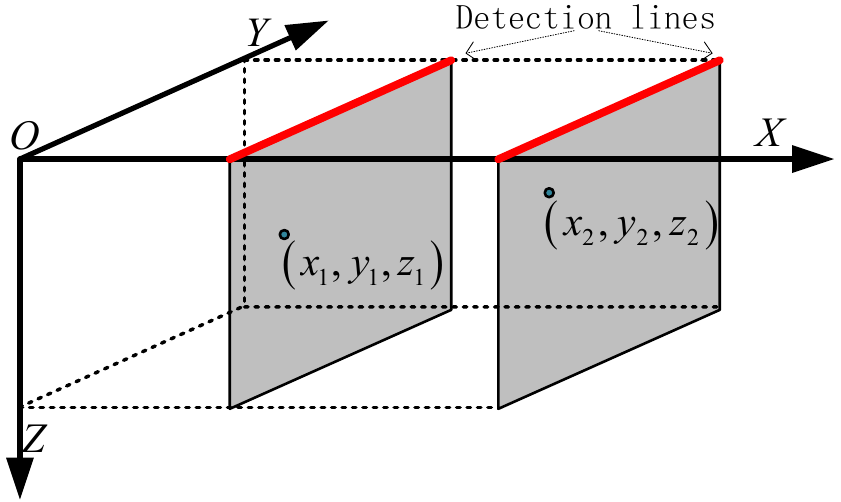}
	\caption{The established coordinate system. The $X,Y,Z$ axises and $O$ are marked. The red line indicates the detection line which is parallel to each other and also parallel to $Y$ axis. }
	\label{coordinate}
\end{figure}

The conducted coordinate system is visualized in Fig. \ref{coordinate}, where the $Y$ axes indicates the direction where GPR is moved (parallel to the detection line), and the direction of $X$ is perpendicular to $Y$. The downward direction perpendicular to the horizontal $XOY$ plane is set to be the positive direction of $Z$ axes. Multiple parallel detection lines are established (the red line in Fig. \ref{coordinate}), where the GPR is moved to obtain B-scan images (a gray surface in Fig. \ref{coordinate} represents a GPR B-scan image).   

%In this paper, the origin of the coordinate system of the detected area is the southwest corner, and the east and north directions are the positive directions of two perpendicular axes $X$ and $Y$ on the ground. In the vertical direction, the downward direction perpendicular to the horizontal plane is set to be the positive direction of $Z$ axes. As Fig. \ref{coordinate} shows, 

The GPR B-scan image obtained at each detection line is processed by the model introduced in Section \uppercase\expandafter{\romannumeral2}. If a hyperbolic point cluster generated by a cable is identified and fitted, the top of the hyperbola is a detected point on the cable\cite{jiang2019cable}, and the coordinates on $Y$ and $Z$ of this point are recorded. The coordinate on $X$ axis is obtained from distance between the detection line to the origin $O$. Based on the above, the cable's location in a detection line could be expressed by a detected point $\left ( x_{i},y_{i},z_{i} \right )$ that is made of the coordinates on the three axises, and the cable's location obtained from all the detection lines would compose a point set as
\begin{equation}
	\textbf{P}=\left \{ P_i\left ( x_{i},y_{i},z_{i} \right ) |\ 0\leq i\leq n\right \}
\end{equation}
in which $\left ( x_{i},y_{i},z_{i} \right )$ indicates the coordinates and $n$ is the number of detected points.
It should be noted that when $i \neq j$, $x_i$ might be equal to $x_j$, since there could be more than one detected cables in a GPR B-scan image obtained at a detection line. As the detection line gradually moves away from the origin $O$, the coordinates on $X$ axis of the acquired detected point on each detection line are increasing, which means for $i < j$, there is $x_i \leq x_j$.

Based on the conducted coordinate system, the location of a buried cable could be described by two respective functions that map any real number $x \in [x_0,x_n]$ on the $X$ axis to $y$ and $z$ on the $Y$ and $Z$ axises. 
The location $\textbf{L}$ of the buried cables in the detected area could be described as:
\begin{equation}
	\begin{aligned}
		\textbf{L}=\{  (\textit{f}_{yc}(x),\textit{f}_{zc}(x),\sigma_c(x) ) \vert \ &c\in [1, C], c\in \mathbb{N},\\ &x\in [x_0,x_n], x\in \mathbb{R} \}
	\end{aligned}	
\end{equation}
where $C$ is the number of cables in the detected area, $\textit{f}_{yc}(x)$ and $\textit{f}_{zc}(x)$ indicate the two functions of the $c$th cable, which map any real number $x \in [x_0,x_n]$ on the $X$ axis to corresponding $y$ and $z$ on the $Y$ and $Z$ axis. $\sigma_c(x)$ is the confidence interval at any $x \in [x_0,x_n]$, which indicates the minimum interval of each $x$ where the pipeline existence probability is greater than or equal to $95\%$.

\subsection{Clustering Detected Point into Different Cables}
The proposed GPR B-scan image interpreting model could estimate more than one cables in an GPR B-scan image, thus detected points from several cables could be obtained at a detection line.
As the bending of the underground cables should be limited, otherwise the cable might be damaged\cite{jiang2019cable}, thus the detected point ($x_{i},y_{i},z_{i}$) of the cable at $l$th detection line could be selected by:
\begin{equation}
	\label{chooses} 
	\min
	(\frac{\vec D_{i}\cdot \vec D_{j}}{|\vec D_{i}|\times|\vec D_{j}|}),
\end{equation}
where $\vec{D_{i}} = (y_{i}-y_{j},z_{i}-z_{j})$, and $y_{j}$, $z_{j}$ are from the points on the $(l-1)$th detection line. For the first detection line, the cable's direction could be assumed to be perpendicular with the detection line, thus $(x_{-1},y_{-1},z_{-1})=(-1,y_{0},z_{0})$.

Equations \eqref{chooses} is applied to process the detected point from the detection line closest to the origin $O$, line by line along the $X$ axis. Each cable could generate only one detected point in a detection line. If a detected point is obtained in a detection line that does not belong to any previous cable, a new cable will be created and the direction of this cable is set to be perpendicular with the detection line. When all the detected points are processed, the cable with too short length are discarded, since there might be underground objects incorrectly identified as the buried cable  (for example, when the the distance between two nearest detection lines is about one meter, the cable segment that contains only one or two detected points should be discarded). After that, the partition of $\textbf{P}$ is created as 
\begin{equation}
	\label{pointset}
	\{\textbf{P}_c| c=1,2,\cdots,C\},
\end{equation}
in which $\textbf{P}_c$ indicates the set of detected points generated from the $c$th cable.

\subsection{Cable Fitting Algorithm Based on Gaussian-Process Regression}

Two kinds of algorithm are usually adopted in regression or fitting. The first one requires a pre-defined function, and the parameters of this function are then adjusted to fit the given data. 
The second kind of algorithms takes all kinds of functions into account and chooses the one which is the most consistent with the data and could achieve the maximum likelihood. 
Considering the existing noises of depth and position when detecting buried cables, using a pre-defined function would lead to inaccurate results, thus applying function with the maximum likelihood is more tolerant in this situation. In this paper, the Gaussian-process regression is expanded in fitting detected points on the cable to obtain the curve function with the maximum likelihood.
% and calculate the variance to indicate the position of buried cable.

As Equation \eqref{pointset}, $\textbf{P}_c$ indicates a set of detected points generated by a cable, which is a subset of $\textbf{P}$. 
It is assumed that the horizontal location $y_i$ and the vertical location $z_i$ are independent from each other but both related to $x_i$, thus $\textbf{P}_c$ could be separated into two data subsets:
\begin{equation}
	\textbf{P}_{c1}=\left\{(x_i, y_i)|0 \leq i \leq n\right\},
\end{equation}
\begin{equation}
	\textbf{P}_{c2}=\left\{(x_i, z_i)|0 \leq i \leq n\right\}.
\end{equation}
These two two-dimensional subsets $\textbf{P}_{c1}$ and $\textbf{P}_{c2}$ are fitted separately, and then integrated into the location information of the entire cable in three-dimensional space.

A Gaussian process indexed by $\boldsymbol{x}=x_0,x_1,\cdots,x_n$ (time or space) is a stochastic process such that every finite collection of random variables has a multivariate normal distribution, and it could be specified by its mean function $\textit{m}(\boldsymbol{x})$ and covariance function $\textit{k}_{f}(x_i, x_j)$, ($0\leq i,j \leq n$).
Intuitively, a real valued function $\textit{f}(\cdot)$ could be viewed as a vector with infinite dimensionality, and could be sampled from a normal distribution which specifies the function space.
Therefore, Gaussian processes define the prior distribution of a latent function $\textit{f}(\cdot)$, and could encode the assumptions of $\textit{f}(\cdot)$ in the design of covariance function $\textit{k}_{f}(\cdot, \cdot)$ instead of choosing any specific form of $\textit{f}(\cdot)$.

For the point set $\textbf{P}_{c1}=\left\{(x_i, y_i)|0 \leq i \leq n\right\}$, the Gaussian-process regression model could be described as:
\begin{equation}
	y_i = \textit{f}(x_i) + \epsilon_i,\ \  i=0,1,...,n,
\end{equation}
where $\epsilon_i$ are noise variables with independent normal distribution, then a priority distribution over functions $\textit{f}(\cdot)$ could be assumed:
\begin{equation}
	\textit{f}(\boldsymbol{x})\sim\mathcal{GP}(m(\boldsymbol{x}),\boldsymbol{K}(\boldsymbol{x}, \boldsymbol{x})),
\end{equation}
where $\boldsymbol{x} = {(x_i)}_{i=1}^n$,  $\textit{m}(\cdot)$ is the mean function, and $\boldsymbol{K}(\cdot, \cdot))$ is covariance matrix satisfying $[\boldsymbol{K}(\boldsymbol{x}, \boldsymbol{x})]_{ij}$=$\textit{k}_{f}(x_i, x_j)$.
The mean function $\textit{m}(\cdot)$ could be assumed as zero to reduce the amount of parameters\cite{williams2006gaussian}. 
The covariance function $\textit{k}_{f}(x_i, x_j)$ maps the distance between $x_i$ and $x_j$ to the covariance between $\textit{f}(x_i)$ and $\textit{f}(x_j)$, which forms as: 
\begin{equation}
	\label{conv}
	k_f(x_i, x_j) = \exp{(-\frac{1}{2\beta^2}(x_i - x_j)^2)}+\theta^2\delta_{ij},
\end{equation}
where $\beta$ is a hyperparameter, and $\delta_{ij}$ is a Kronecker delta which is one iff $i=j$ and zero otherwise.
The covariance function indicates that the closer two detected points are, the higher correlation they have. This is justified since the actual location of target point on the cable is more related to the nearby points, and Equation \eqref{conv} reduces the noises from the detected points in the distance. In the case of the same distance, the smaller the $\beta$, the larger the covariance, which means that the cable position at this point is more correlated with the surrounding cable positions. In actual engineering, $\beta$ is affected by the material of the cable. The greater the bendability of the buried cable material, the smaller the correlation between a point on the cable and the surrounding points, since this cable could be arbitrarily bent or change direction. Considering the existence of noises in detecting, the term $\theta^2\delta_{ij}$ is added to the function to follow the independence assumption about the noise from the positioning method or the depth information obtained by interpreting B-scan images.

To calculate the location $y^*$ for a new position $x^*$, it is assumed that $y^*$ and all regression targets $\boldsymbol{y}$ in $\textbf{P}_{c1}$ has the same joint normal distribution with zero mean function, where $\boldsymbol{y} = {(y_i)}_{i=1}^n$. Then we have:

\begin{equation}
	\left( \begin{matrix}\boldsymbol{y} \\ y^* \end{matrix}\right) \sim \mathcal{N}\left(\left(\begin{matrix}\boldsymbol{0}\\0\end{matrix}\right), \left(\begin{matrix}\boldsymbol{K}&\boldsymbol{k}^*\\{\boldsymbol{k}^*}^T&k^{**}\end{matrix}\right)\right),
	\label{jointDis}
\end{equation}
where $\boldsymbol{k}^*$ is a covariance vector in which the $i$th element is $[\boldsymbol{k}^*]_{i}$=$\textit{k}_{f}(x^*, x_i)$ and $k^{**} = 1$, and $\boldsymbol{k}^{*T}$ is the transposed matrix of $\boldsymbol{k}^{*}$. Then the distribution of $y^*$ could be calculated via Equation \eqref{jointDis} as:
\begin{equation}
	y^* \sim \mathcal{N}\left(\mu^*, \sigma^*\right).
\end{equation}
The mean value $\mu^*$ and variance value $\sigma^*$ of target $y^*$ could be calculated through Equations \eqref{mu} and \eqref{sigma} as:
\begin{equation}
	\mu^* = {\boldsymbol{k}^*}^T\boldsymbol{K}^{-1}\boldsymbol{y},
	\label{mu}
\end{equation}
\begin{equation}
	\sigma^* = -{\boldsymbol{k}^*}^T\boldsymbol{K}^{-1}\boldsymbol{k}^* + k^{**}
	\label{sigma}
\end{equation}
where $\mu^*$ indicates the mean value of the output of all valid functions in which these functions fit the detected data points, and $\mu^*$ serves as the most likely location of the buried cable.
Considering the existence of noises in the detected points, $\sigma^*$ presents the confidence interval by $\mu^*\pm 2\sigma^*$.

The above process is applied to both $\textbf{P}_{c1}$ and $\textbf{P}_{c2}$ separately to obtain the location on $Y$ and $Z$ axis, and then combined to describe the three dimensional location of the buried cable. For each cable, given an $x$ in the detected area, the corresponding $y$ and $z$ could be obtained. When mapping the buried cable, the noise of $y$ mainly comes from the positioning error, and the main source of the noise of $z$ is the error produced from interpreting the GPR B-scan image to obtain the cable's depth. The noises of $y$ and $z$ could be regarded as independent and unrelated, and the hyperparameter $\theta$ in the processing of $\textbf{P}_{c1}$ and $\textbf{P}_{c2}$ are decided separately as $\theta_y$ and $\theta_z$, which depends on the intensity of the two kinds of noises.

\subsection{The Pseudo Code of Locating Buried Cables}

The pseudo code of clustering and fitting the obtained detected point set is presented as Algorithm \ref{algo}.

\begin{algorithm}[t]
	\small
	\caption{\small{Locating buried cables by Gaussian-process regression}}
	\label{algo}
	{\bf Input:} 
	Detected points set $\textbf{P}=\left\{(x_i, y_i, z_i)|0 \leq i \leq n\right\}$, hyperparameters $\beta$ and $\theta$.\\
    {\bf Output:}
	The most likely locations and confidence intervals of all buried cables $	\textbf{L}=\{  (\textit{f}_{yc}(x),\textit{f}_{zc}(x),\sigma_c(x) ) \}$.
	\begin{algorithmic}[1]
		\State Cluster the points in $\textbf{P}$ into different independent subset$\{\textbf{P}_c| c=1,2,\cdots,C\}$ via Equation \eqref{chooses}, where $n_c$ indicates the number of detected points in $\textbf{P}_c$.
		\For{every $\textbf{P}_c$ in $\textbf{P}$}
		\State Separate $\textbf{P}_c$ into $\textbf{P}_{c1} = (\boldsymbol{x}_c, \boldsymbol{y}_c) = \left\{(x_{ci}, y_{ci})|0 \leq i \leq n_c\right\}$ and $\textbf{P}_{c2} = (\boldsymbol{x}_c, \boldsymbol{z}_c) = \left\{(x_{ci}, z_{ci})|0 \leq i \leq n_c\right\}$,
		\State Calculate the covariance matrix $\boldsymbol{K}_c$ on $\boldsymbol{x}_c$, where $[\boldsymbol{K}_c(\boldsymbol{x}_c, \boldsymbol{x}_c))]_{ij}$=$\textit{k}_{f}(x_{ci}, x_{cj})$, and
		\begin{equation}\nonumber
			k_f(x_{ci}, x_{cj}) = \exp{(-\frac{1}{2\beta^2}(x_{ci} - x_{cj})^2)} + \theta^2\delta(x_{ci}, x_{cj}).
		\end{equation}
		\For{every $x^*_{ci} \in [x_{c0},x_{cn_c}]$}
		\For{every $x_{cj} \in \boldsymbol{x}_c$}
		\State Calculate the covariance between $x^*_{ci}$ and $x_{cj}$:
		\begin{equation}\nonumber
			k^*_{cj} = \exp{(-\frac{1}{2\beta^2}(x^*_{ci} - x_{cj})^2)}.
		\end{equation}
		\EndFor
		\State $\boldsymbol{k}^*_c = \left\{k^*_{cj}|x_{cj} \in \boldsymbol{x}_c \right\}$,
		\State The function of the most likely locations $\textit{f}_{yc}$ and $\textit{f}_{zc}$:
		\begin{equation}\nonumber
			\textit{f}_{yc}(x^*_{ci}) = {\boldsymbol{k}_c^*}^T\boldsymbol{K}_c^{-1}\boldsymbol{y},
		\end{equation}
		\begin{equation}\nonumber
			\textit{f}_{zc}(x^*_{ci}) = {\boldsymbol{k}_c^*}^T\boldsymbol{K}_c^{-1}\boldsymbol{z},
		\end{equation}
		\State The function of the confidence intervals $\sigma_c$:
		\begin{equation}\nonumber
		\sigma_c(x^*_{ci}) = -{\boldsymbol{k}^*}^T\boldsymbol{K}^{-1}\boldsymbol{k}^* + 1,
		\end{equation}
		\begin{equation}\nonumber
		\textit{f}_{yc}(x^*_{ci})\pm\sigma_c(x^*_{ci}),\ \  \textit{f}_{zc}(x^*_{ci})\pm\sigma_c(x^*_{ci}).
		\end{equation}
	
		\EndFor
		\EndFor
		\State \Return  The most likely locations and confidence intervals of buried cables $	\textbf{L}=\{  (\textit{f}_{yc}(x),\textit{f}_{zc}(x),\sigma_c(x) ) \vert c\in [1, C], c\in \mathbb{N}, x\in [x_0,x_n], x\in \mathbb{R} \}$.
	\end{algorithmic}
\end{algorithm}

\section{Experimental Study}
\label{experiments}

In this section, experiments on real-world datasets are conducted. After that, the analysis of the experimental results and some comparative work are presented.

\subsection{Experimental Environment and Settings}
\begin{figure}[htbp]
	\centering
	\subfigure[]{ \centering
		\label{area1}
		\includegraphics[height=1.11in]{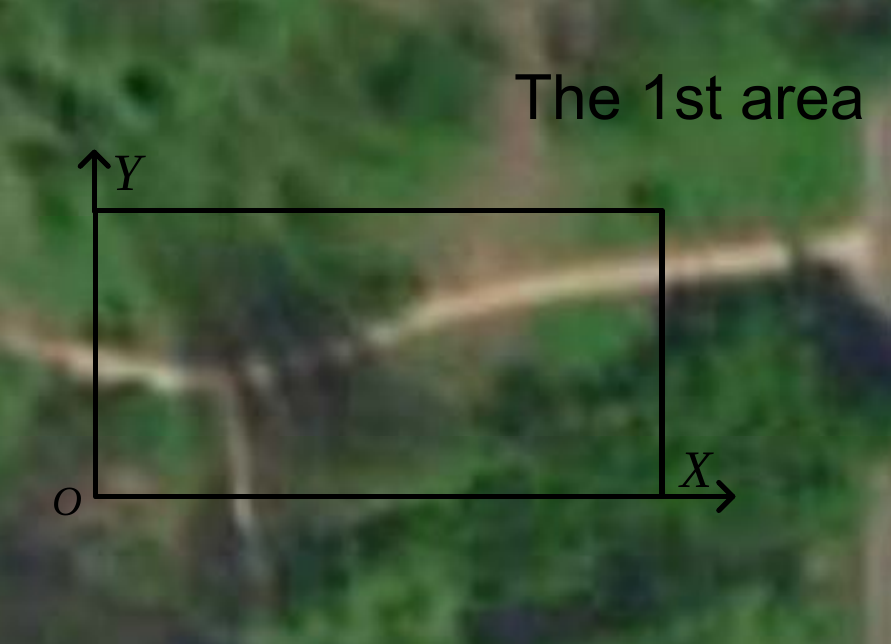}}
	\subfigure[]{ \centering
		\label{area2}
		\includegraphics[height=1.11in]{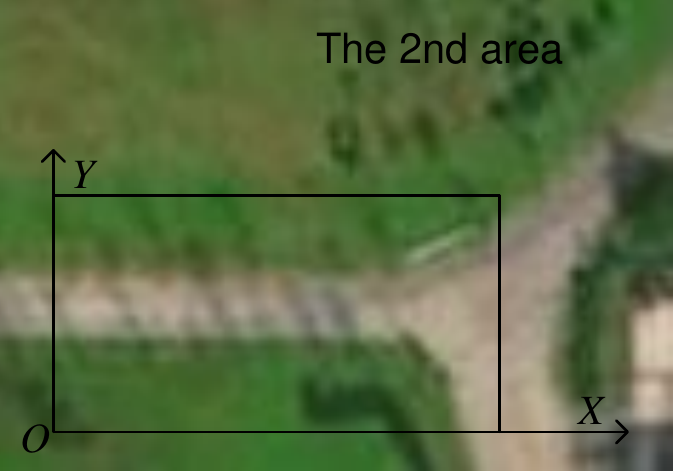}}
	\subfigure[]{ \centering
		\label{area3}
		\includegraphics[height=1.03in]{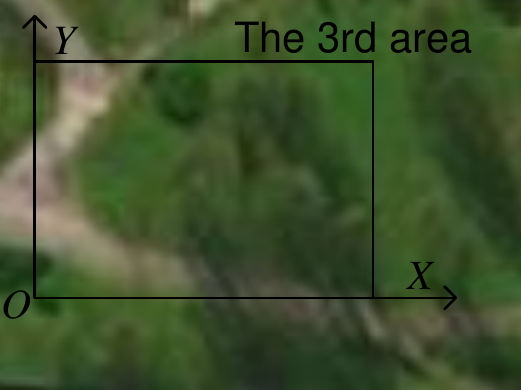}}
	\subfigure[]{ \centering
		\label{gssi2}
		\includegraphics[height=1.03in]{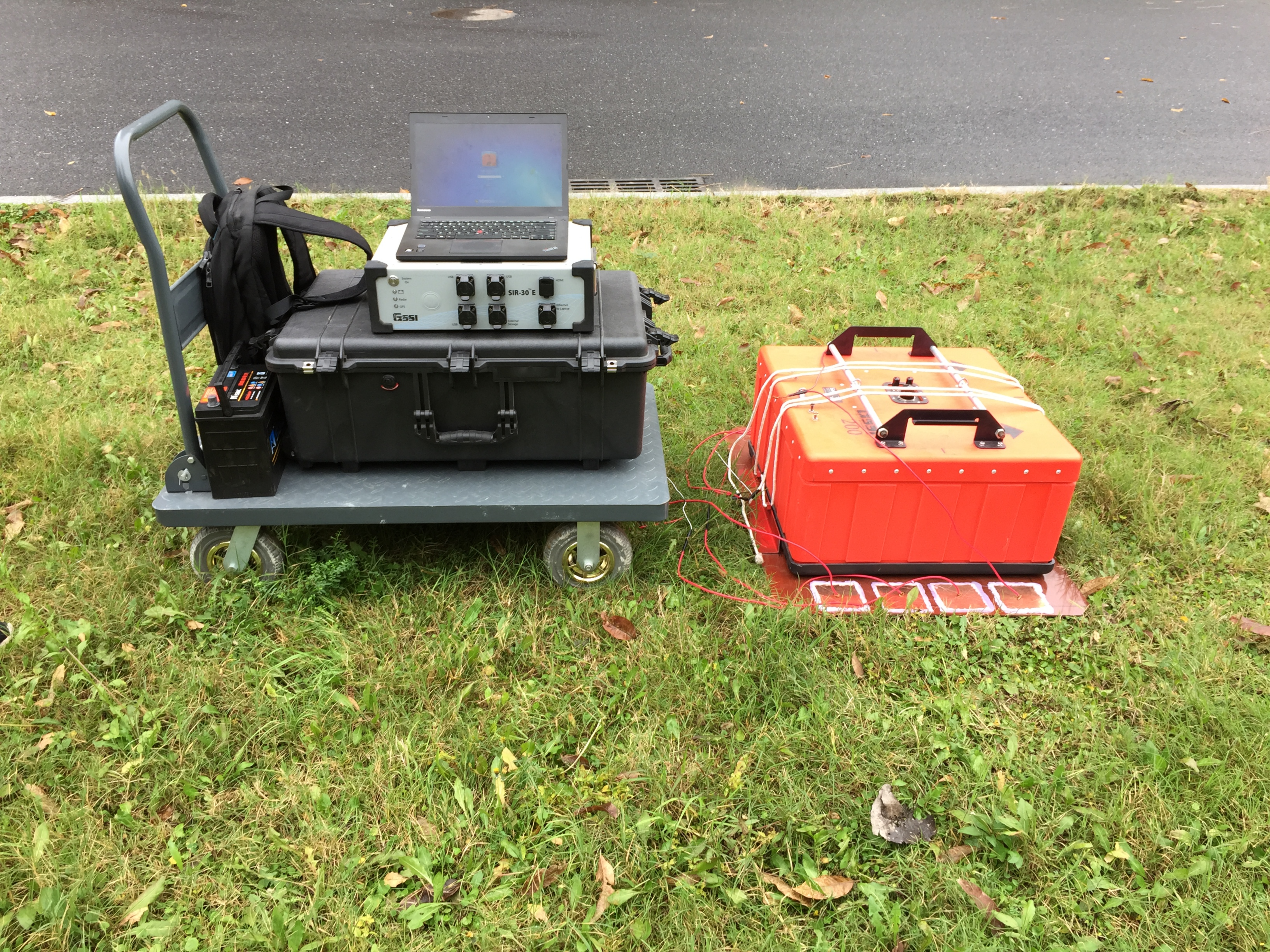}}
	\caption{The three selected areas and the established coordinate systems are shown as (a), (b) and (c). The GSSI's SIR-30 GPR with 200MHz antenna is adopted in our experiments to obtain the B-scan image at each detection line as (d). (e) and (f) show the utilized GPR hast and antenna.}	\label{area_and_device}
\end{figure}

\begin{figure}[htbp]
	\centering
	\subfigure[]{ \centering
		\label{cable1}
		\includegraphics[height=1.1in]{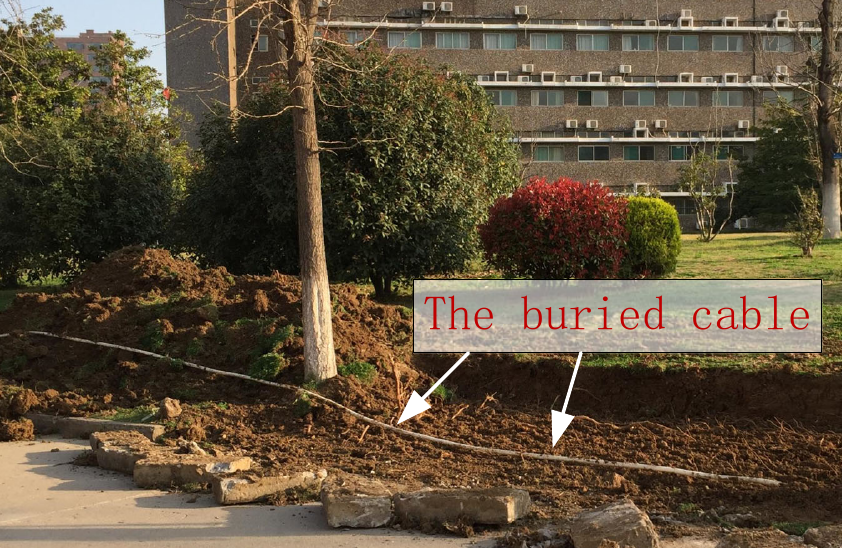}}
	\subfigure[]{ \centering
		\label{cable2}
		\includegraphics[height=1.1in]{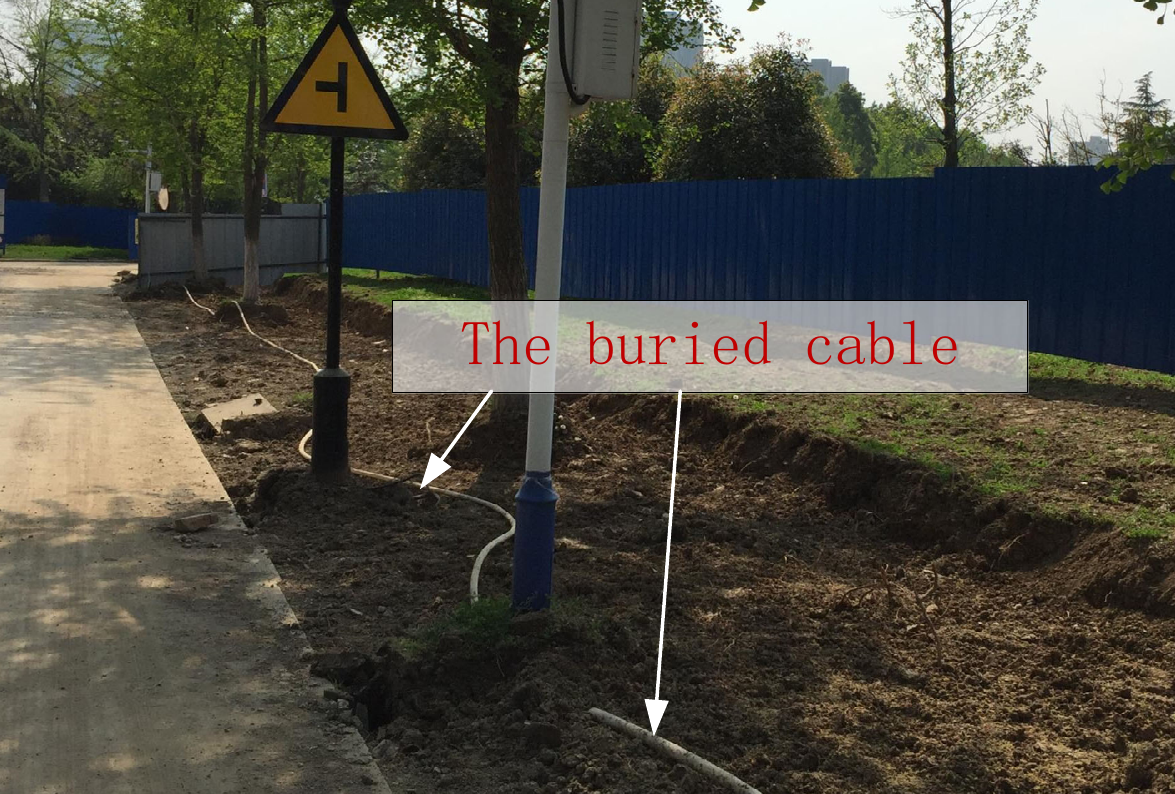}}
	\caption{The monitoring probes around the experimental area. It could be seen that some buried cables have been excavated, as shown as the white line. The two figures also demonstrate that the buried cables could not be described by straight lines.}\label{cable12}
\end{figure}

By consulting the existing underground pipeline map, three experimental areas are identified. The existing piping map in these areas provide the start and end points of each section of cables, which are connected by straight lines.
GSSI’s SIR-30 GPR with 200-MHz antenna is utilized to collect GPR B-scan images, and the GPR's supporting positioning equipment would record the position of every point along the detected path. When a hyperbola is identified and fitted, the position of the buried cable at this point could be obtained and recorded. The three selected areas, the established coordinate systems and the utilized GPR and antenna are visualized in Fig. \ref{area_and_device}. These areas are all near the roads with electric utilities nearby, such as cameras, street lights, etc. Part of the buried cables in these areas are evacuated as Fig. \ref{cable12}, which demonstrates the fact that underground cables could not be accurately located by straight line segments.


When conducting the coordinate system, we chose the direction of the detection line to make it as perpendicular as possible to the direction of the cables on the existing pipeline map. The dimensions of the first and second detection areas are both $20$m in length ($X$ axis) and $10$m in width ($Y$ axis). The size of the third detection area is $20$m long ($X$ axis) and $15$m wide ($Y$ axis). In these three areas, detection lines are conducted parallel to each other and also parallel to the $Y$ axis every $2$m. For the hyperparameters, $\beta$ is set to be $1$, and $\theta_y=0.3$, $\theta_z=0.1$, since the error of the utilized GPR's supporting positioning equipment could be controlled within $0.3$m, while the depths of all detected cables are less than $1$m, and the change of the depth of each cable is also less than $0.1$m, acknowledged from the existing pipeline map.

\subsection{Experimental Results}

\begin{figure*}[htbp]
	\centering
	\subfigure[]{ \centering
		\label{fit11}
		\includegraphics[height=1.24in]{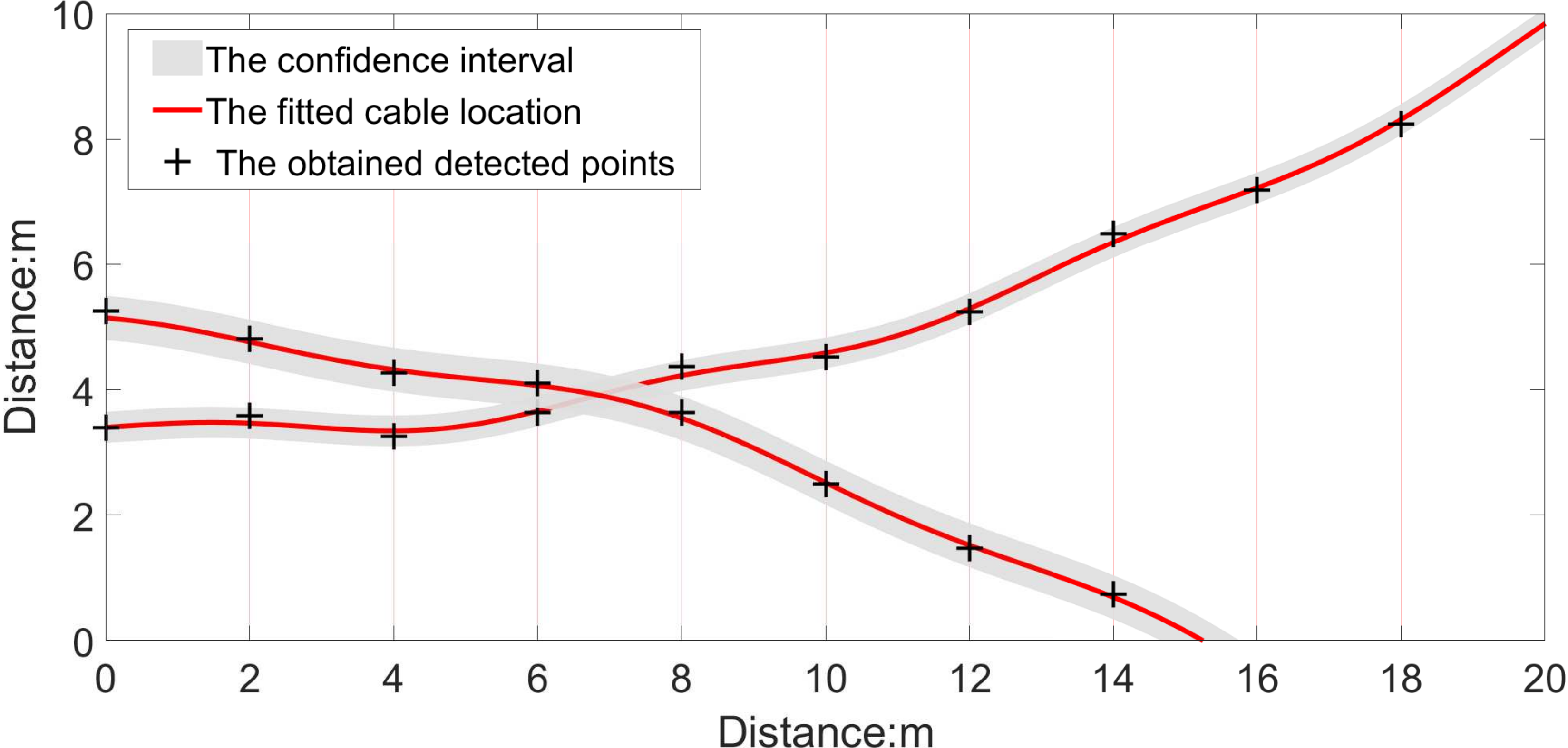}}
	\subfigure[]{ \centering
		\label{fit21}
		\includegraphics[height=1.24in]{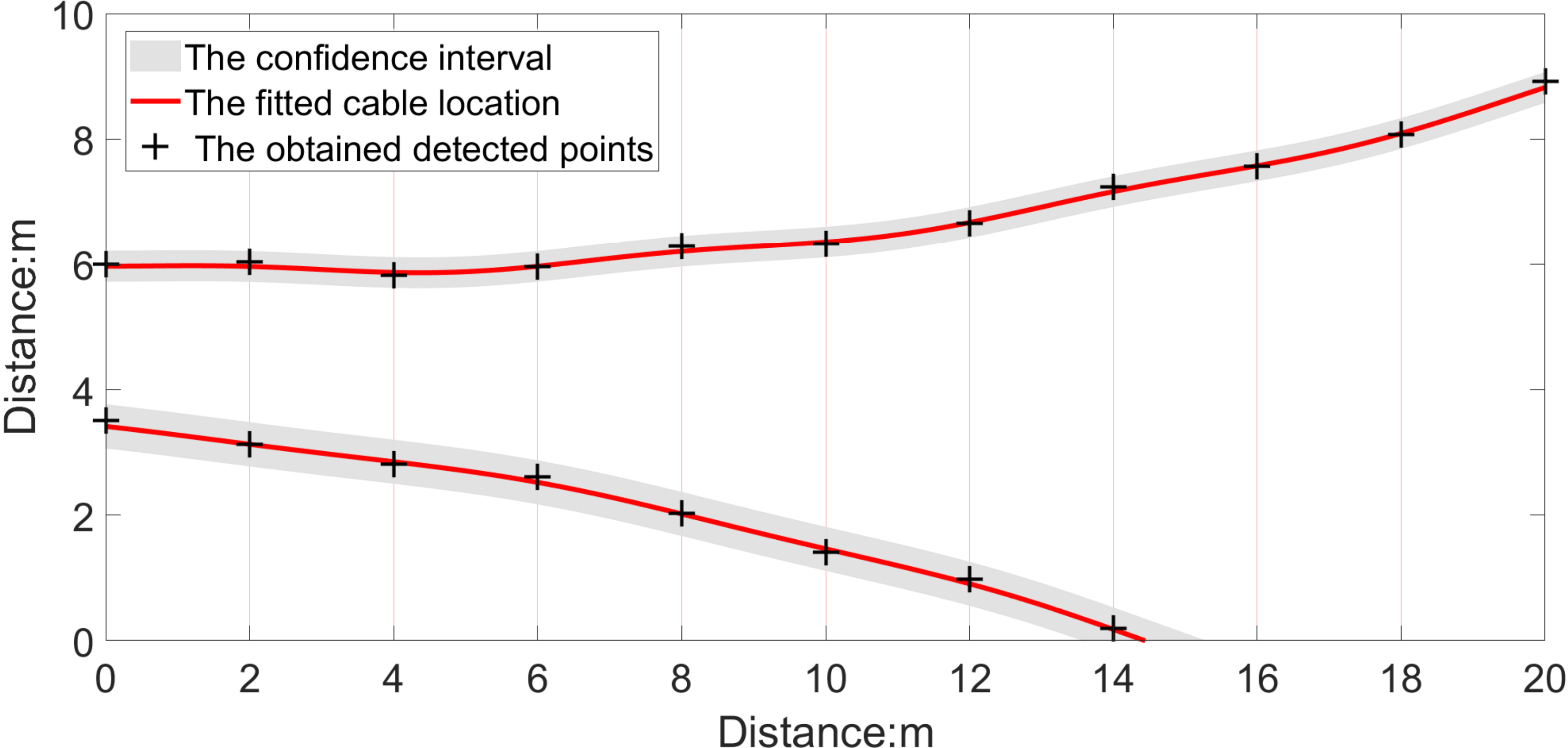}}
	\subfigure[]{ \centering
		\label{fit31}
		\includegraphics[height=1.24in]{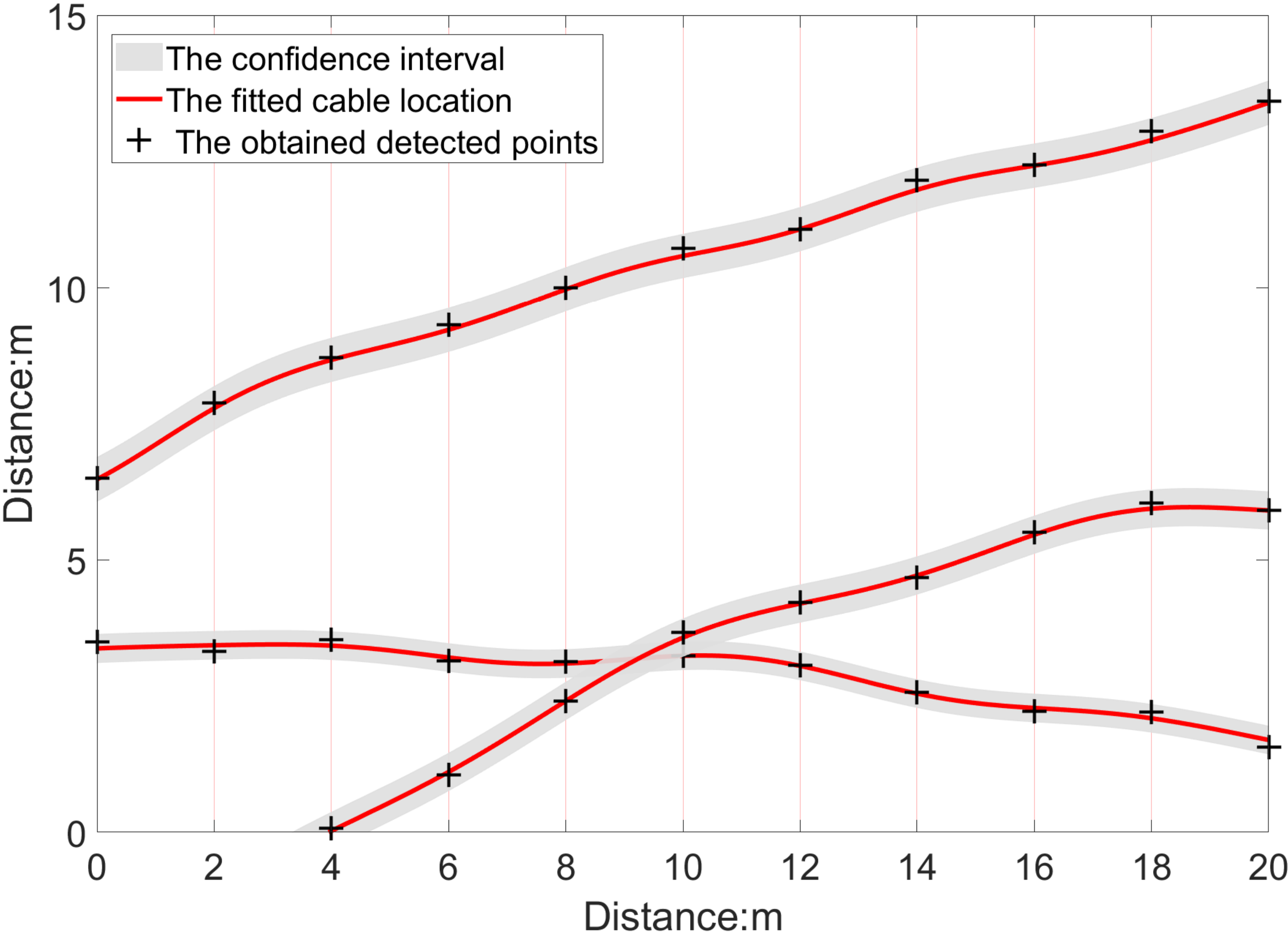}}
	\caption{The fitting results on $XOY$ plane. The vertical line perpendicular to the $X$ axis in the coordinate system represents the established detection line. It could be seen that the cable fitted by the proposed algorithm does not pass through the location of each detected point, since the proposed algorithm takes the noise during positioning into account. At the same time, the output location of each cable is a curve, which is consistent with the actual situation. The confidence intervals are also obtained.}\label{fitcable12}
\end{figure*}

\begin{figure}[htbp]
	\centering
	\subfigure[]{ \centering
		\label{Bscan31}
		\includegraphics[width=0.23\textwidth]{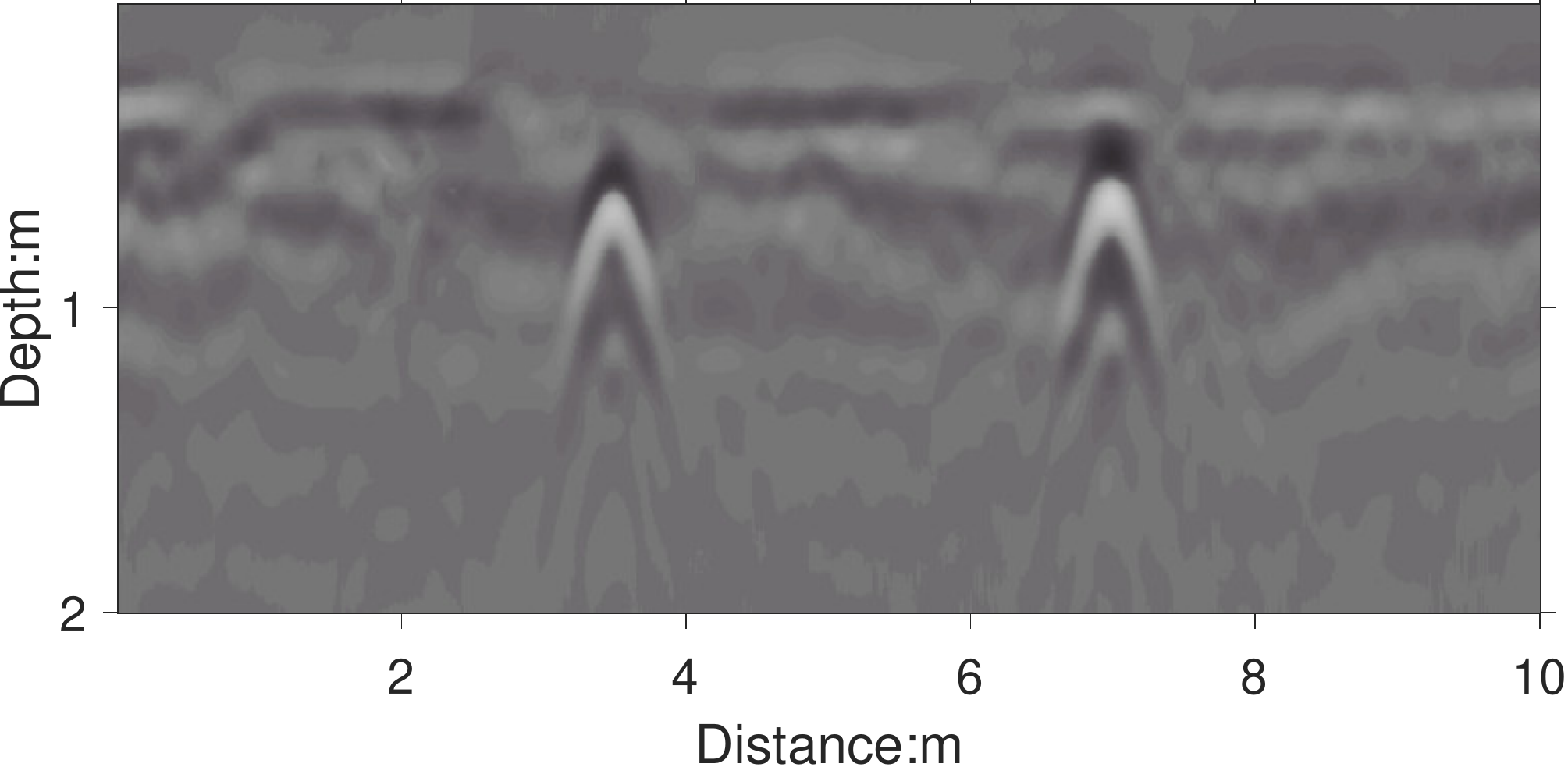}}
	\subfigure[]{ \centering
		\label{Bscan32}
		\includegraphics[width=0.23\textwidth]{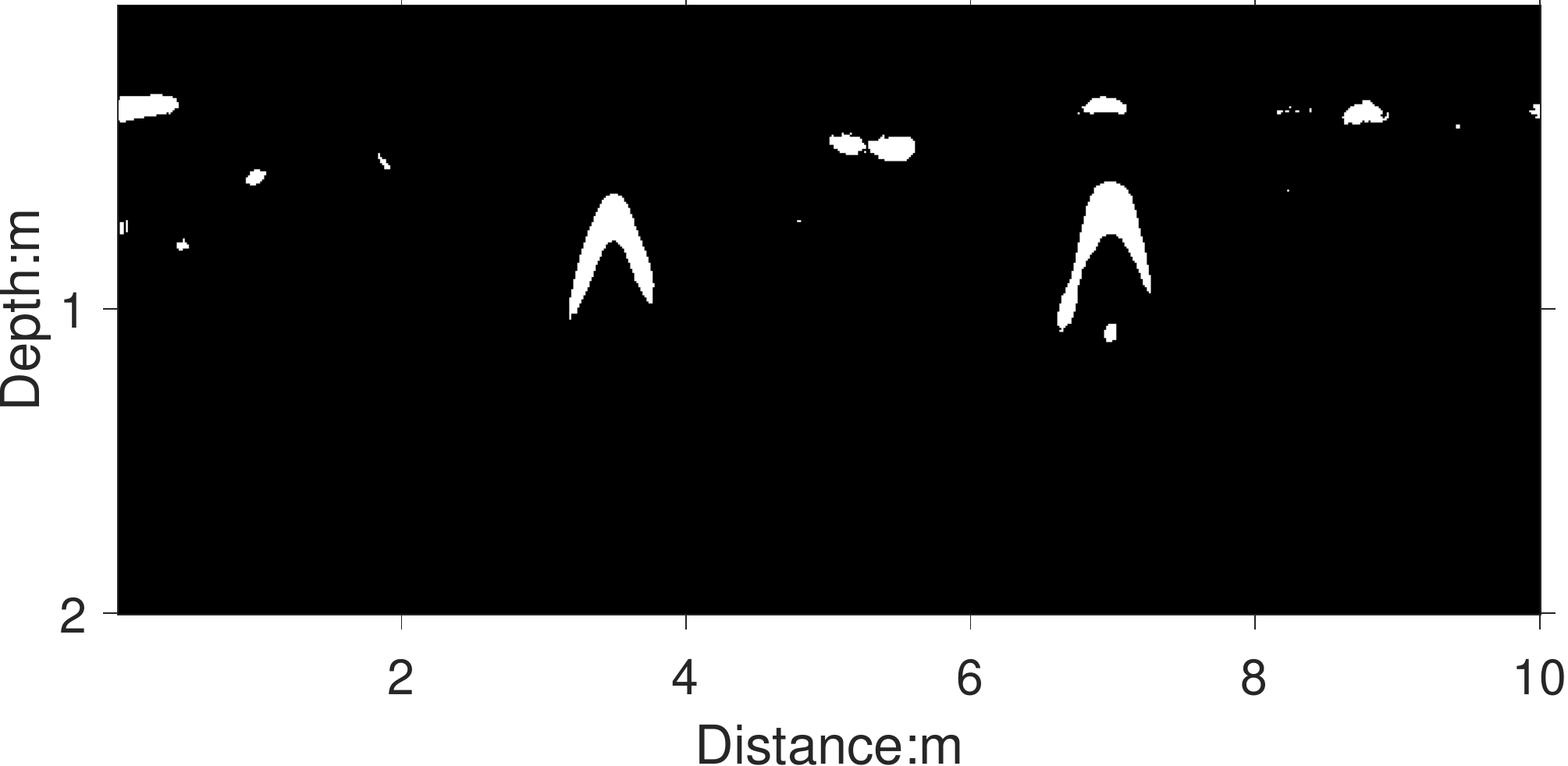}}
	\subfigure[]{ \centering
		\label{Bscan33}
		\includegraphics[width=0.23\textwidth]{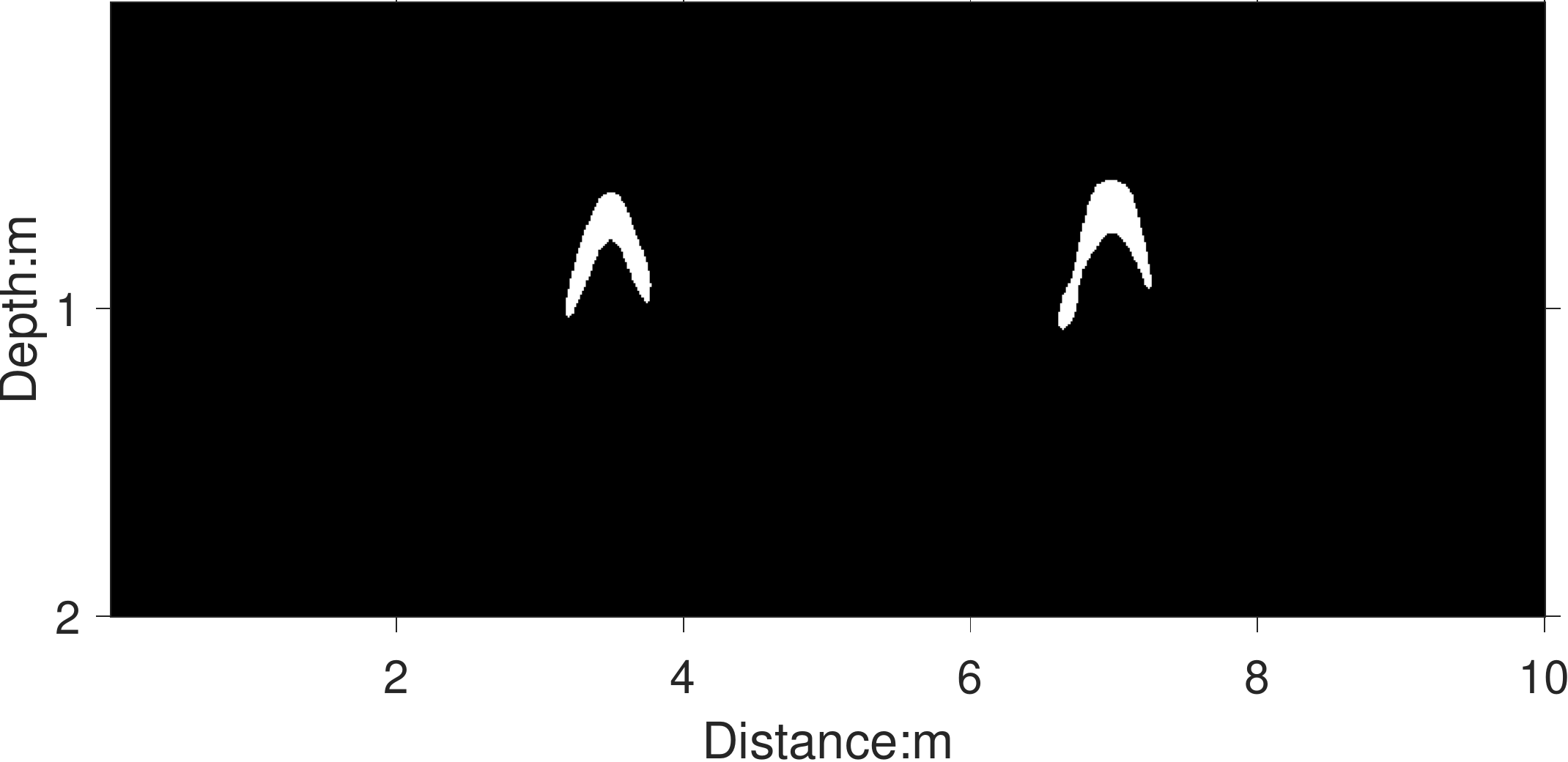}}
	\subfigure[]{ \centering
		\label{Bscan34}
		\includegraphics[width=0.23\textwidth]{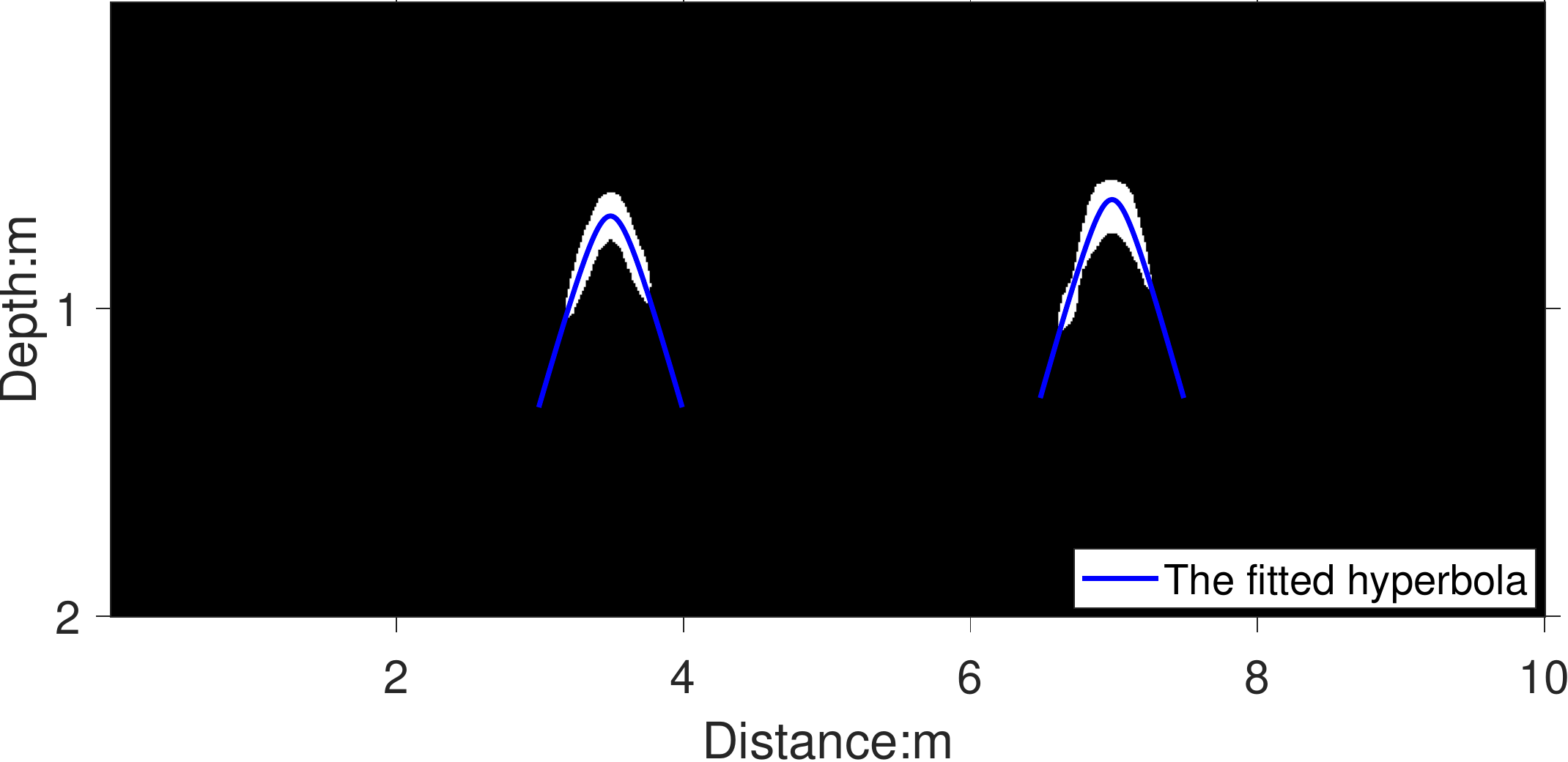}}
	\caption{The processing flow of interpreting a GPR B-scan image. (a) is the original image. (b) is the preprocessed binary imaged. (c) is the obtained result after OSCA and PFJ with only two regions, of which the fitting results are shown in (d).}\label{bscanpro}
\end{figure}

\begin{figure}[htbp]
	\centering
	\subfigure[]{ \centering
		\label{Bscan21}
		\includegraphics[width=0.23\textwidth]{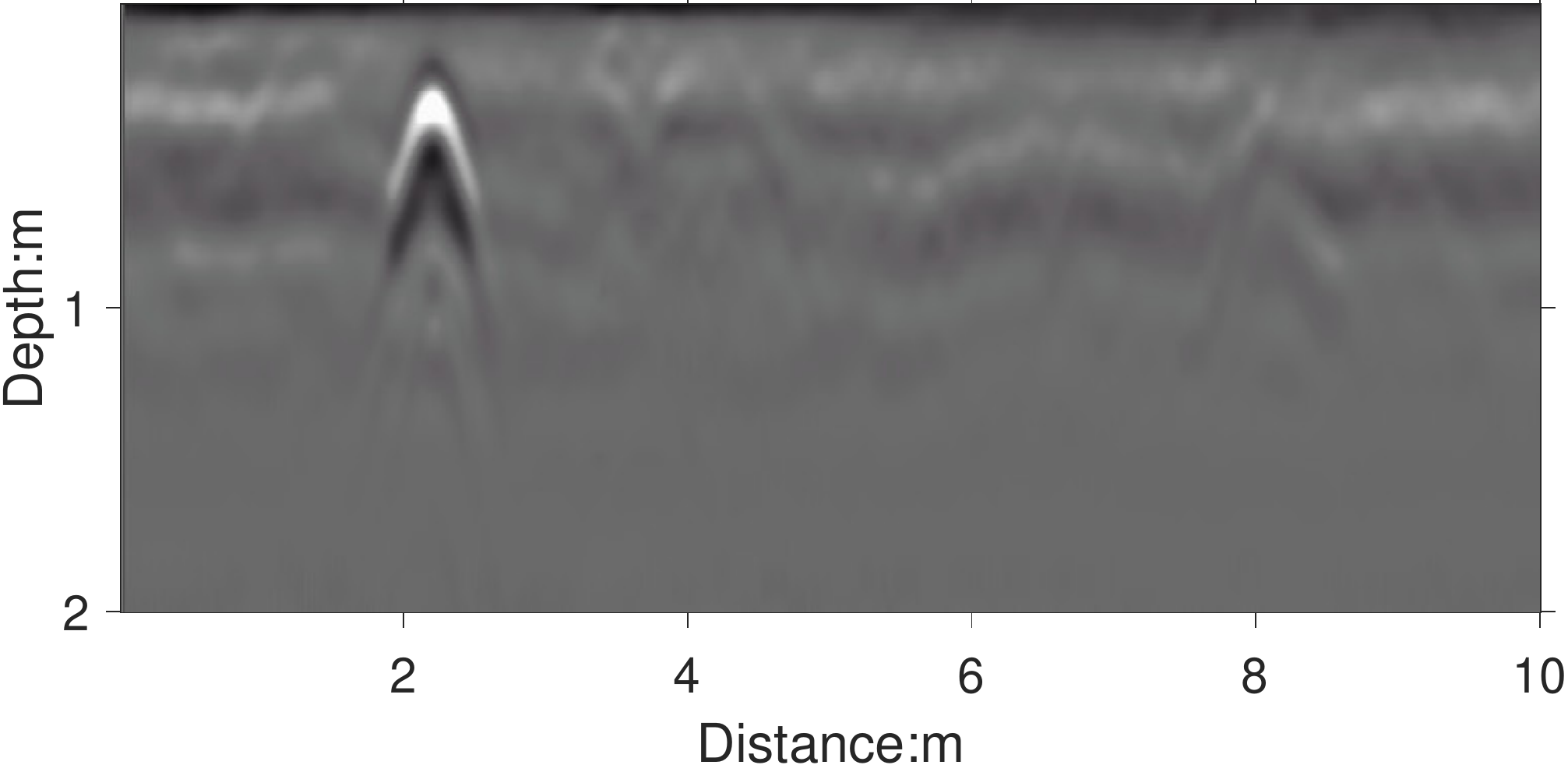}}
	\subfigure[]{ \centering
		\label{Bscan11}
		\includegraphics[width=0.23\textwidth]{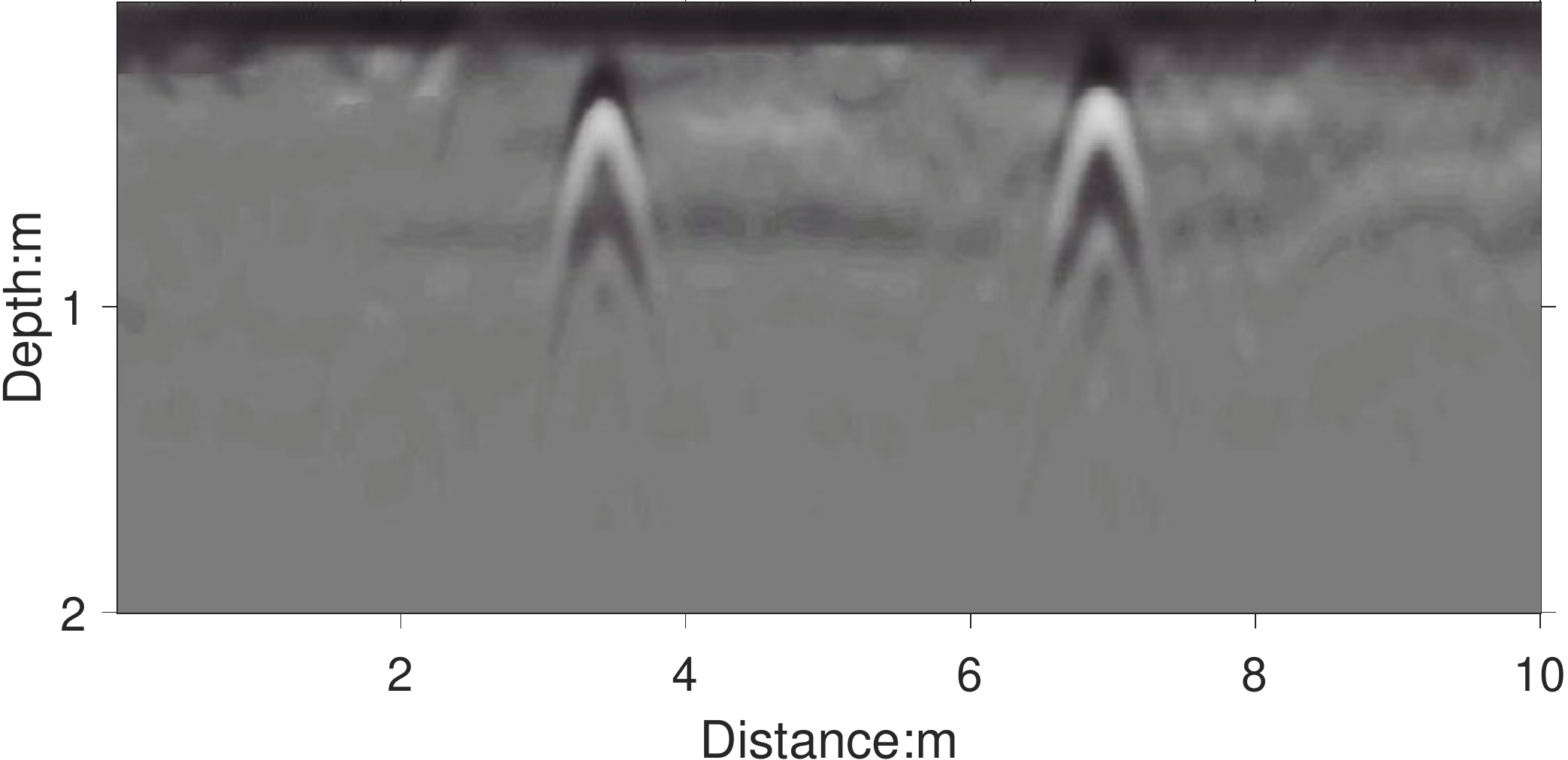}}
	\subfigure[]{ \centering
		\label{Bscan22}
		\includegraphics[width=0.23\textwidth]{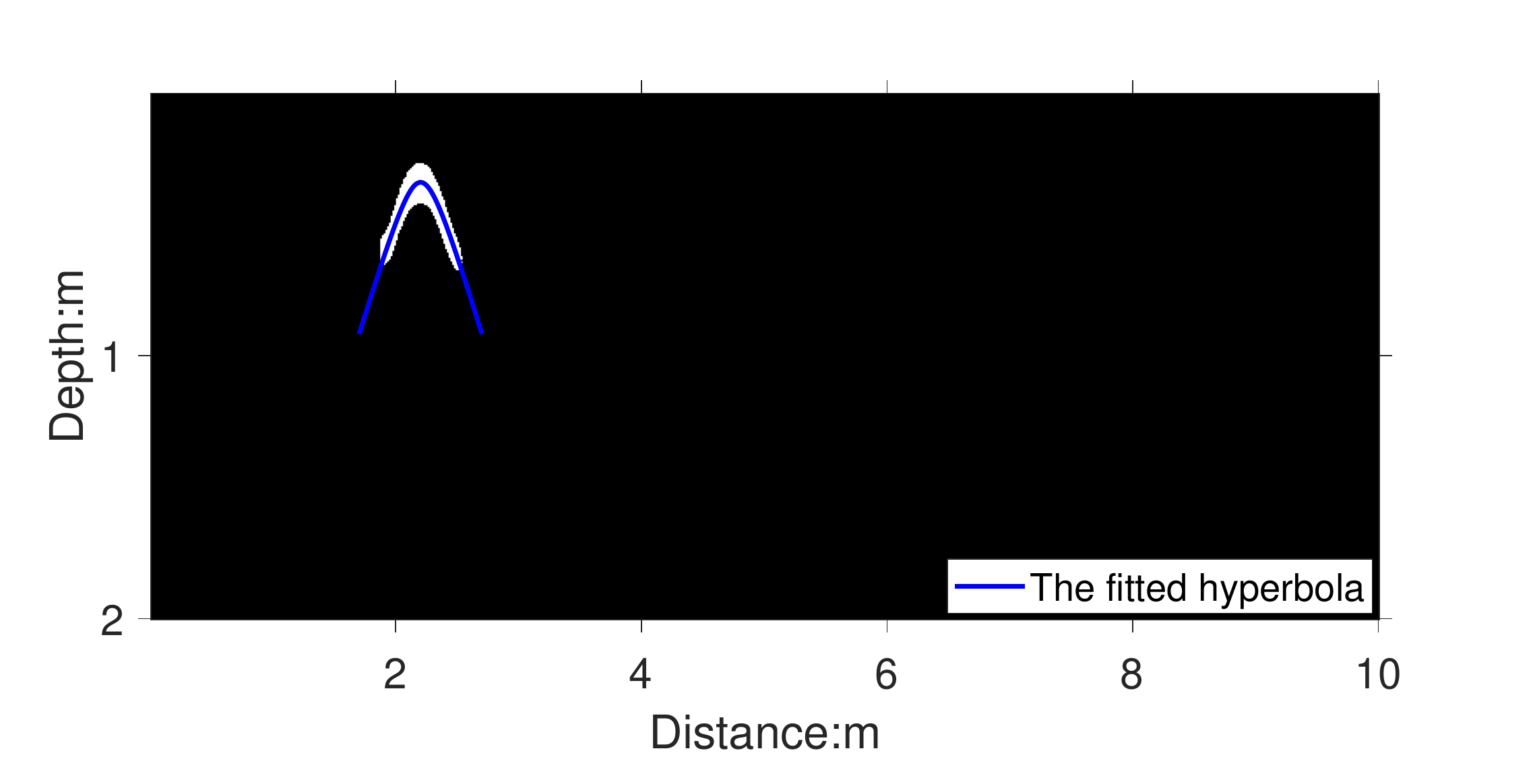}}
	\subfigure[]{ \centering
		\label{Bscan12}
		\includegraphics[width=0.23\textwidth]{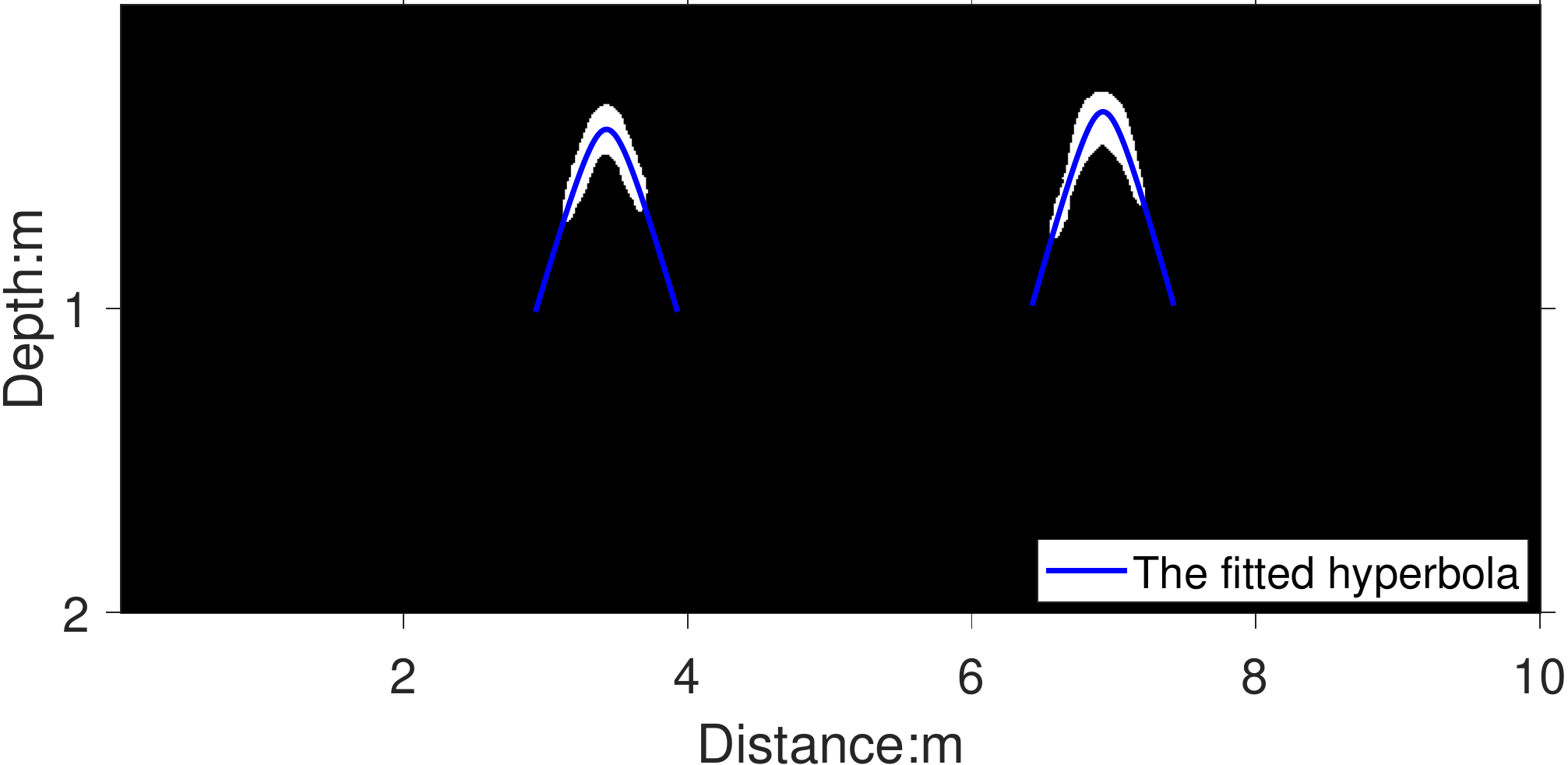}}
	\caption{Two obtained GPR B-scan images and the fitting results. (a) and (b) are the two original images, and (c) and (d) are the fitting results.}\label{bscanexample}
\end{figure}

In our experiments, the depths of buried cables are less than $1$m with changes less than $0.1$m, while the movements of cables are not straightforward. Thus the $XOY$ plane of the established coordinate system is adopted to visualize the cable fitting results as Fig. \ref{fitcable12}, that is, the same perspective as the satellite map from top to bottom. The detected points at each detection line are obtained by interpreting the GPR B-scan image utilizing the proposed GPR B-scan image interpreting model. Due to the limitation of the length of this paper, these images could not be fully demonstrated. Fig. \ref{bscanpro} illustrates the process of interpreting a GPR B-scan image, and part of obtained images with processing results are shown in Fig. \ref{bscanexample}.

\begin{comment}
\begin{figure}[htbp]
	\centering
	\includegraphics[width=0.43\textwidth]{fit31}
	\caption{The obtained fitting result on $XOY$ plane of the third areas. The vertical line perpendicular to the $X$ axis in the coordinate system represents the established detection line}
	\label{fitcable3}
\end{figure}
\end{comment}

The accuracy of the proposed model could be measured by computing the average error $E$ of depth and horizontal location, which could be calculated as
\begin{equation}
	\label{Error}
	\small
	E =\frac{1}{m} \sum_{k=1}^m|calculated\_value_k - measured\_value_k|
\end{equation}
where the $measured\_value$ comes from the actual positions and depths of the points on the buried cables at established detection lines and another $20$ randomly selected points on each detected cable. $m$ indicates the number of these measured points. The positions and depths of these points are accurately located in the coordinate system to ensure the accuracy of the $measured\_value$. The $calculated\_value$ comes from the positions and depths obtained by the proposed method. Experimental results are presented in Table \ref{ErrorAVE}. The errors of depths are within $7$cm while the errors of positions are within $12$cm. In addition, the actual positions of all the points in the experiments are within the obtained confidence intervals. In real-world applications, the obtained interval could provide early warning for excavation work, and it could also reduce the detection range for precise detection between detection lines. The specific analysis of the experimental results and comparisons with other methods are presented in the follows.
\begin{table}[htbp]
	\scriptsize
	\centering
	\caption{Errors of proposed model}
	\begin{tabular}{ccccccc}
		\toprule
		\multirow{3}[6]{*}{Area} & \multicolumn{6}{c}{The  average error (cm)} \\
		\cmidrule{2-7}          & \multicolumn{2}{c}{Detection line} & \multicolumn{2}{c}{ Randomly selected points} & \multicolumn{2}{c}{Altogether } \\
		\cmidrule{2-7}          & depth & position & depth & position & depth & position \\
		\midrule
		1st & 4.12  & 9.42  & 5.22  & 10.51 & 4.85  & 10.14 \\
		\midrule
		2nd & 5.92  & 12.52 & 4.81  & 9.88  & 5.18  & 10.76 \\
		\midrule
		3rd & 5.66  & 11.23 & 7.11  & 9.81  & 6.62  & 10.28 \\
		\bottomrule
	\end{tabular}%
	\label{ErrorAVE}
\end{table}%

\subsection{Analysis and Comparative Work}

\begin{figure}[htbp]
	\centering
	\subfigure[]{ \centering
		\label{hypcompare1}
		\includegraphics[height=1.45in]{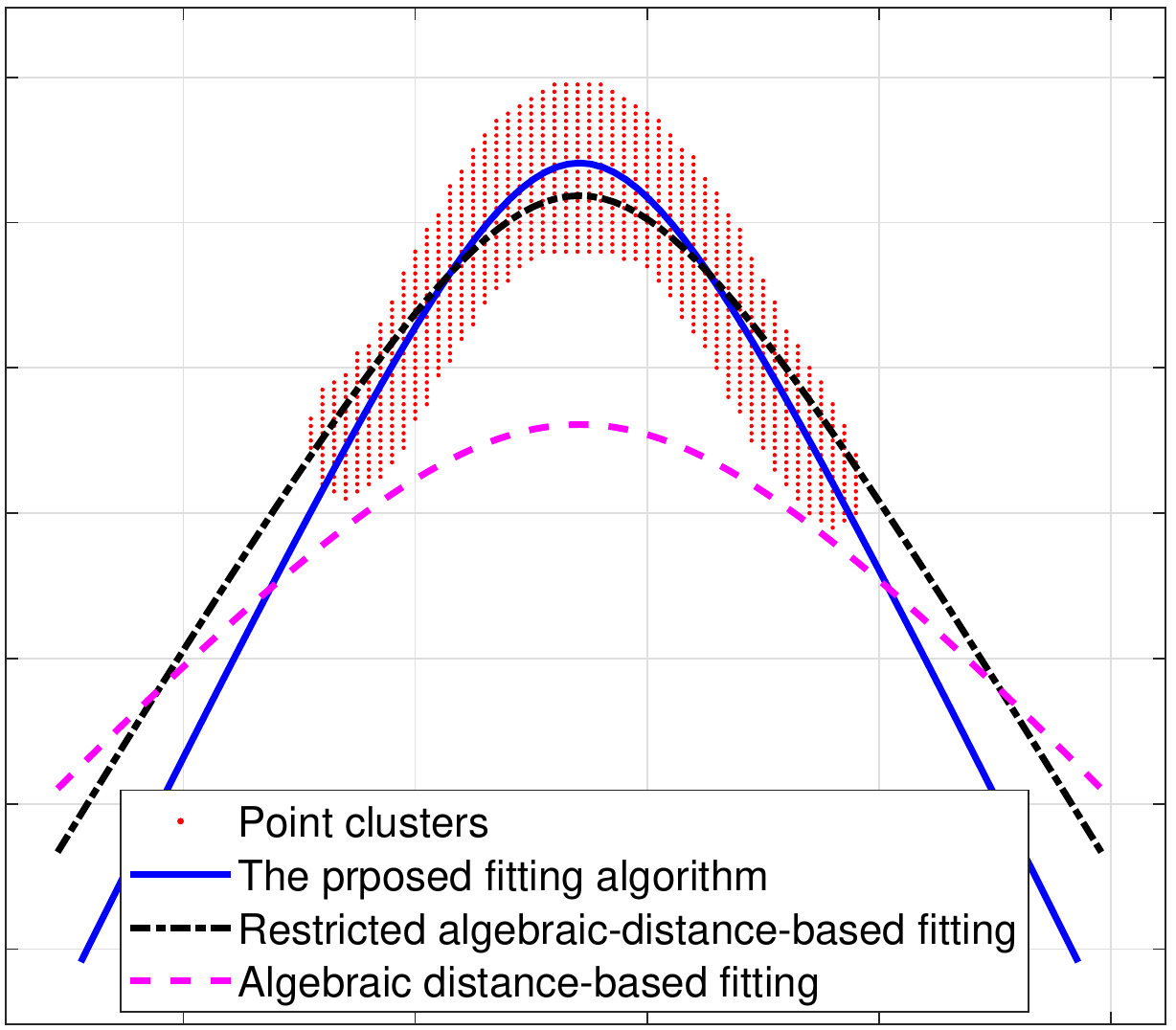}}
	\subfigure[]{ \centering
		\label{hypcompare2}
		\includegraphics[height=1.45in]{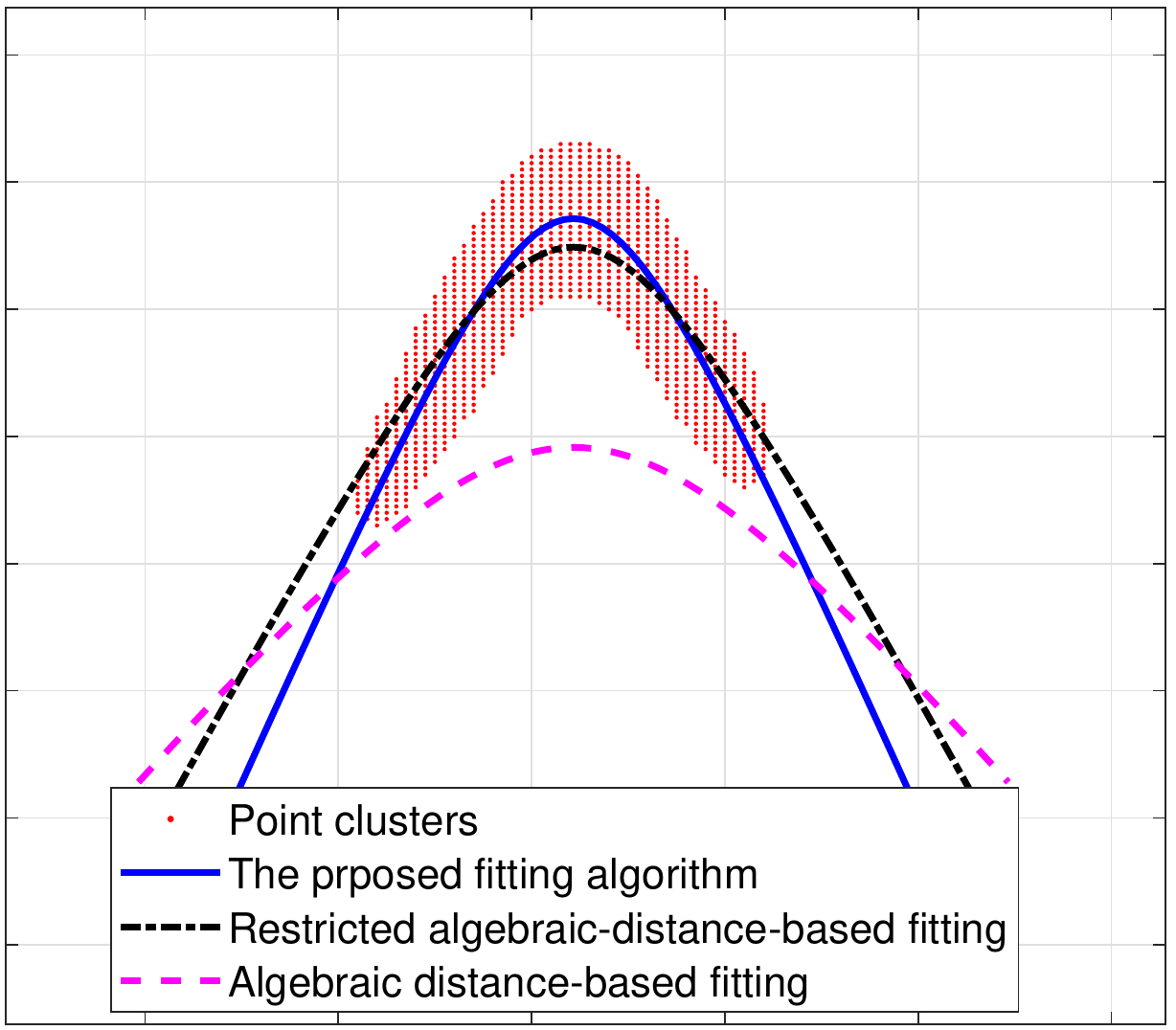}}
	\caption{(a) and (b) show the fitting results of the proposed hyperbolic fitting algorithm, RADF and ADF.}\label{hypcompare}
\end{figure}

In our method, the depth error mainly comes from the process of interpreting GPR B-scan images, especially the hyperbolic fitting. We compared the proposed fitting algorithm with the Restricted Algebraic-Distance-based Fitting algorithm (RADF)\cite{zhou2018automatic} and Algebraic-distance-based Fitting Algorithm (ADF)\cite{chen2010probabilistic}, and Fig. \ref{hypcompare} shows part of the fitting results. The specific average errors of the proposed fitting algorithm, ADF, and RADF are presented in Table \ref{errfitting}.

% Table generated by Excel2LaTeX from sheet 'Sheet2'
\begin{table}[htbp]
	\small
	\centering
	\caption{Errors of proposed hyperbolic fitting algorithm, RADF and ADF}
	\begin{tabular}{cccc}
		\toprule
		\multirow{3}[6]{*}{Area} & \multicolumn{3}{c}{\textbf{The  average error of depth (cm)}} \\
		\cmidrule{2-4}          & The proposed hyperbolic & \multirow{2}[4]{*}{RADF } & \multirow{2}[4]{*}{ADF} \\
		& fitting algorithm &       &  \\
		\midrule
		1st  & 4.85  & 6.11  & 14.39 \\
		2nd   & 5.18  & 6.13  & 17.11 \\
		3rd   & 6.62  & 7.19  & 16.16 \\
		\bottomrule
	\end{tabular}%
	\label{errfitting}%
\end{table}%

When applying ADF, the obtained cable's depth is greater than the actual value in our experiments. Since the hyperbola has two branches (the upper and lower branches), and some points will be fitted to the upper branch of the hyperbola if no restrictions are imposed. RADF improves accuracy by restricting the center of the hyperbola above the point cluster on the basis of ADF. However, when detecting objects with a large difference in permittivity with the surrounding medium, such as buried cables, the response signal could be strong, and the obtained point cluster could be dense. Intuitively, the upper part of the hyperbolic point cluster is relatively thick. In this case, the attraction of the upper half of the hyperbola to the points could not be completely eliminated through RADF. On the basis of RADF, the proposed fitting algorithm further moves the lower part of the hyperbola to the center of the point cluster, thereby improving the fitting accuracy.
And in our experiments, the numbers of iterations of Gauss-Newton iteration in the proposed hyperbolic fitting algorithm are all less than $10$, which verifies the robustness and appropriateness of RADF as the initialization, and also guarantees the efficiency of our model in real-world applications.

The error of positions in our experiments mainly depends on the cable fitting algorithm, and we evaluated the proposed method with the Three-Dimensional Spline Interpolation (TDSI) algorithm\cite{jiang2019cable} and the Marching-Cross-Sections (MCS) algorithm \cite{dou20163d}, which could both obtain the cable's location from the detected points at detection lines. The errors of the three methods are presented in Table \ref{errposition}.
\begin{table}[htbp]
	\small
	\centering
	\caption{Errors of proposed cable fitting algorithm, TDSI and MCS}
	\begin{tabular}{cccc}
		\toprule
		\multirow{3}[6]{*}{Area} & \multicolumn{3}{c}{\textbf{The  average error of position (cm)}} \\
		\cmidrule{2-4}          & The proposed  & \multirow{2}[4]{*}{TDSI } & \multirow{2}[4]{*}{MCS} \\
          & cable fitting algorithm &       &  \\
		\midrule
		1st  & 10.14 & 17.26 & 12.39 \\
		2nd   & 10.76 & 14.19 & 12.21 \\
		3rd   & 10.28 & 15.91 & 14.96 \\
		\bottomrule
	\end{tabular}%
	\label{errposition}%
\end{table}%

The main concern of MCS is that connecting the points on the cable by straight line segments would cause error in the cable's location between connected points. TDSI interpolates the detected points, and locates the buried cable by the interpolated curve. In this process, noises of depth and position at each points are not taken into account. In practice, the errors of TDSI might be greatly affected by the environment and utilized equipment.

\begin{figure}
	\centering
	\subfigure[]{ \centering
		\label{comp1dis1}
		\includegraphics[width=0.34\textwidth]{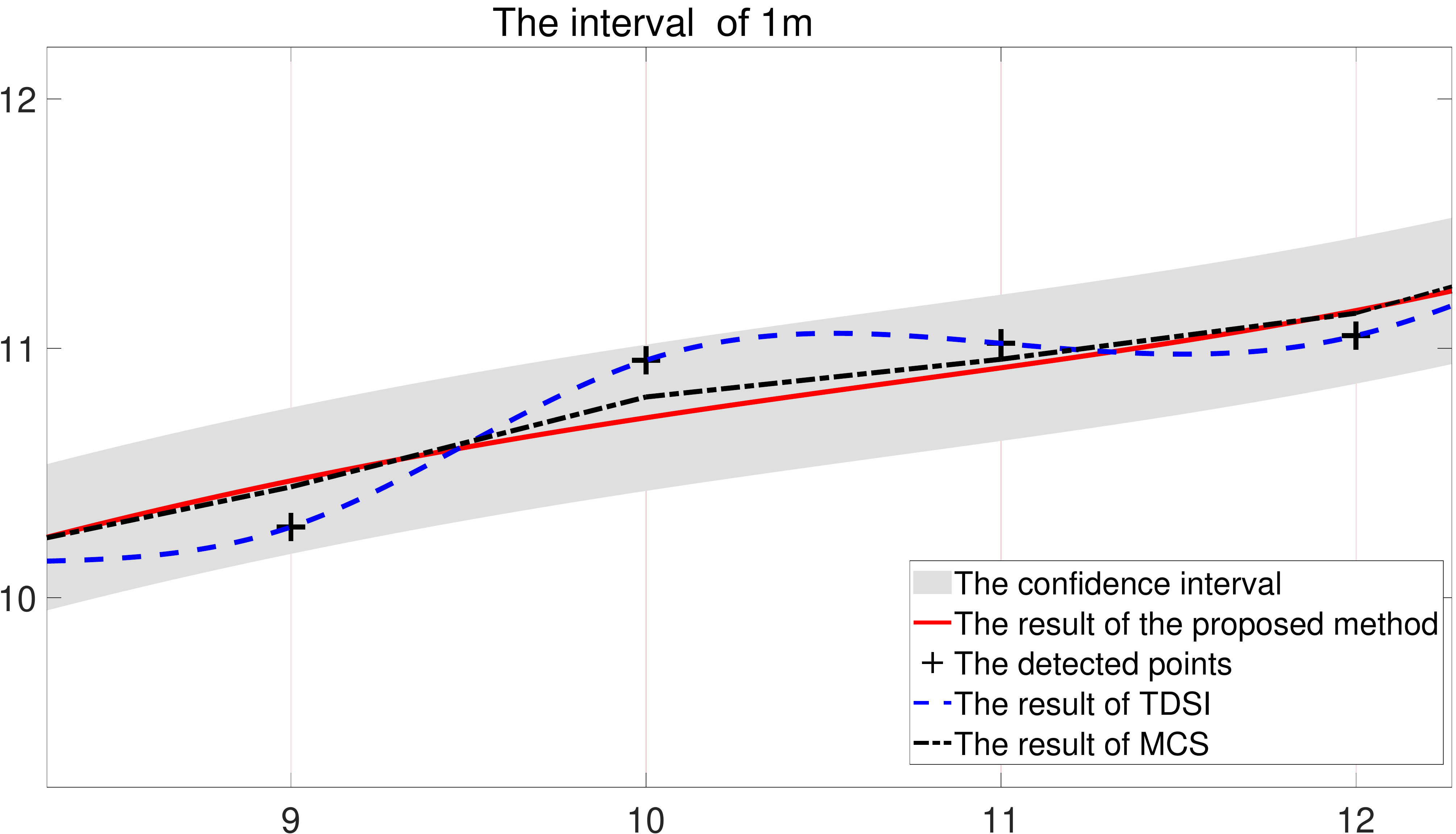}}
	\subfigure[]{ \centering
		\label{comp1dis3}
		\includegraphics[width=0.34\textwidth]{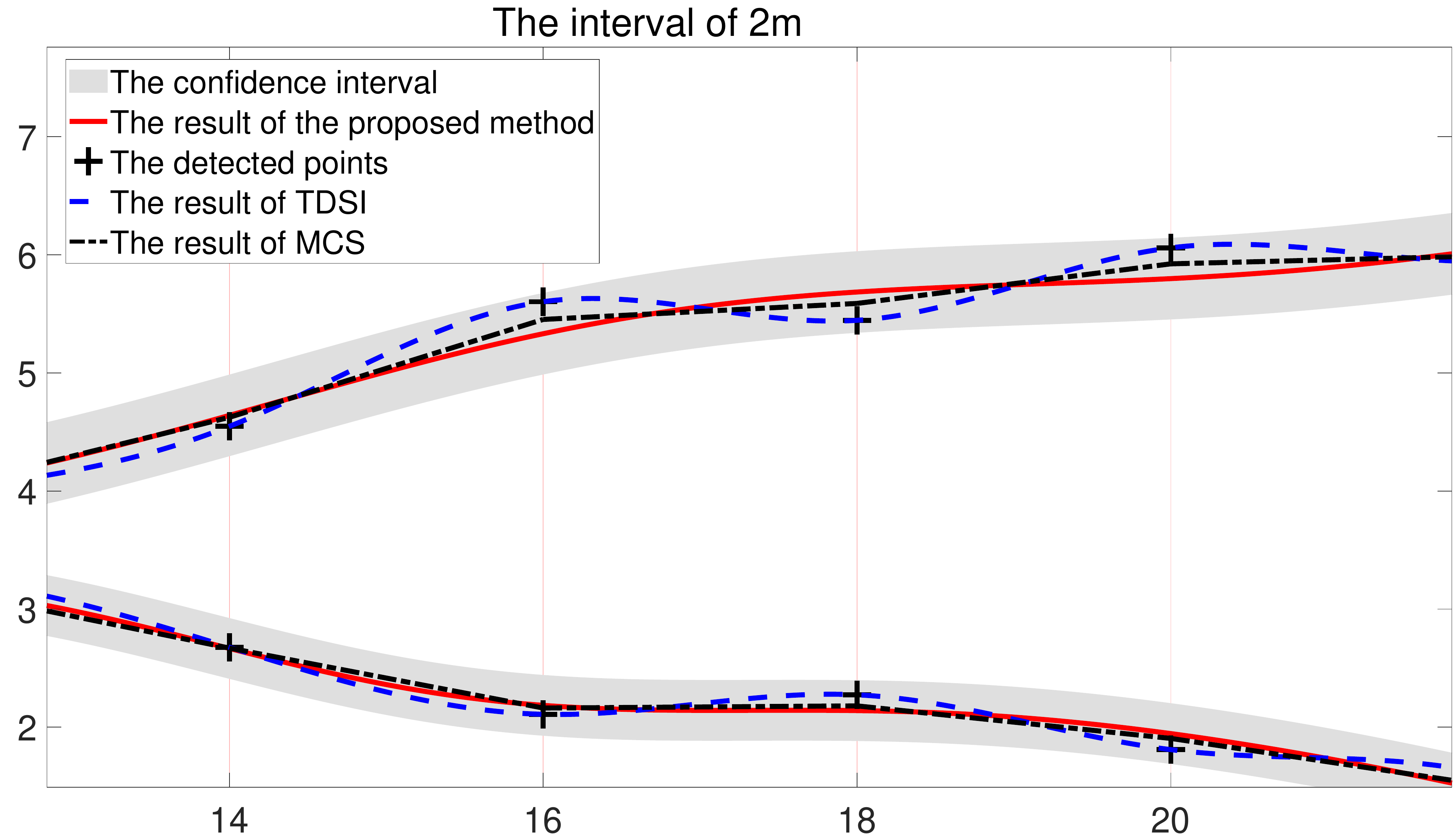}}
	\subfigure[]{ \centering
		\label{comp1dis2}
		\includegraphics[width=0.34\textwidth]{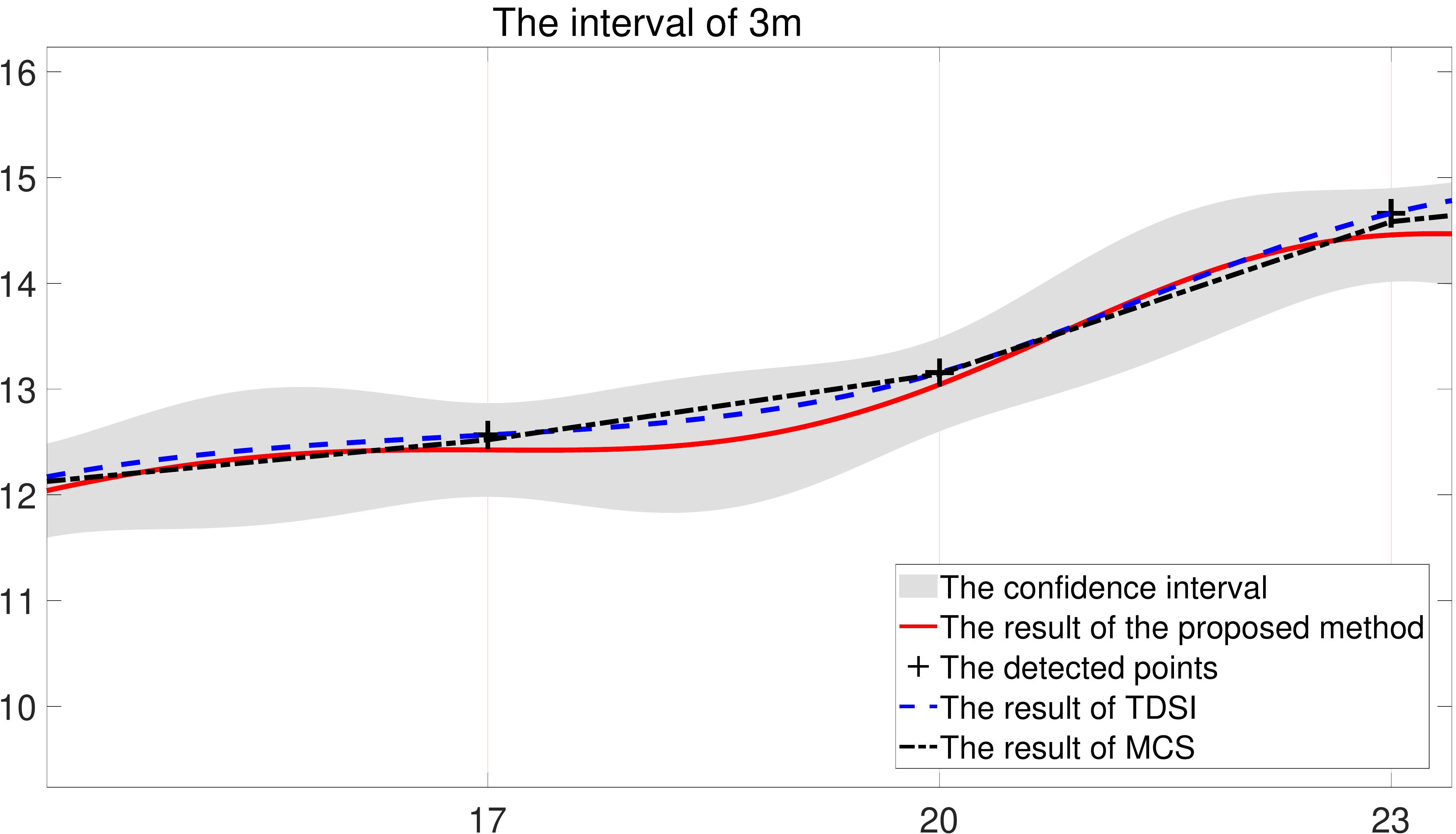}}
	\caption{These three pictures show some details of the results obtained by the proposed cable fitting algorithm, TDSI and MCS in different detection line intervals. As the spacing between detection lines decreases (for example, to $1$m), the results of MCS and the proposed algorithm will gradually approach. As the spacing between detection lines increases (for example, to $3$m), the results of TDSI will gradually approach the results of the proposed algorithm. The proposed algorithm obtains the results closest to the actual situation under different intervals between detection lines, which also proves the robustness of this algorithm.}\label{compdis}
\end{figure}

In order to better evaluate the performance of these three methods, we constructed detection lines with different distance intervals.
Detection lines with intervals of $1$m and $3$m are conducted, and the errors of the three methods are presented in Table \ref{crrpositioncompare}. Due to the limitation of the length of this paper, the results of these experiments could not be fully presented here. The details of some experimental results in different detection line intervals are visualized in Fig. \ref{compdis}.

% Table generated by Excel2LaTeX from sheet 'Sheet4'
\begin{table}[htbp]
	\scriptsize
	\centering
	\caption{Errors of the proposed cable fitting algorithm, TDSI and MCS under different detection line intervals}
	\begin{tabular}{ccccccc}
		\toprule
		\multirow{4}[8]{*}{Area} & \multicolumn{6}{c}{\textbf{The  average error of position (cm)}} \\
		\cmidrule{2-7}          & \multicolumn{3}{c}{1m interval of detection lines } & \multicolumn{3}{c}{3m interval of detection lines  } \\
		\cmidrule{2-7}          & The proposed & \multirow{2}[4]{*}{TDSI } & \multirow{2}[4]{*}{MCS} & The proposed & \multirow{2}[4]{*}{TDSI } & \multirow{2}[4]{*}{MCS} \\         & algorithm &       &       & algorithm &       &  \\
		\midrule
		1st  & 7.64  & 21.46 & 7.65  & 21.14 & 22.18 & 29.49 \\
		\midrule
		2nd   & 7.16  & 21.01 & 8.21  & 19.76 & 24.46 & 31.29 \\
		\midrule
		3rd   & 6.88  & 20.91 & 7.02  & 22.28 & 21.51 & 29.13 \\
		\bottomrule
	\end{tabular}%
	\label{crrpositioncompare}%
\end{table}%

When the distance between the detection lines is $3$m, the errors of the three methods are all larger than that with $2$m intervals. In particular, the error of MCS changes the most. When the distance between the detection lines becomes larger, the length of the straight line segments obtained by MCS gets longer. Thus the details of the cable's location between two connected points could not be well described by MCS. When the interval of detection lines is $1$m, the errors of MCS and the proposed cable fitting algorithm are reduced, while the error of TDSI becomes larger,
since MCS and the proposed cable fitting algorithm consider the existence of noise. When applying these two methods, the greater the density of detected points are, the better the denoising effect and the more accurate the fitting result become. TDSI dose not deal with noise, and when the detected points become dense, the curve obtained by interpolation will get unsmooth with frequently changes of directions.
With intervals of $1$, $2$ and $3$m, the accuracy of the proposed algorithm is stable, which also verifies the robustness of this model in different environments.

\section{Conclusion}

In this paper, a cable locating method based on GPR and Gaussian process regression is proposed. The coordinate system of the detected area is firstly conducted, and the input and output of locating the buried cables are determined. Parallel detection lines are then established, along which the GPR is moved to obtain the B-scan images. After that, hyperbolic shapes on these obtained images are identified and fitted, thus the positions and depths of some points on the cables could be roughly derived. Finally, these points are clustered and fitted to infer the most likely location of the buried cables. Furthermore, the confidence intervals of cables are also obtained by the proposed method, and in the conducted experiments, the actual positions of the cables we inspected are all within the intervals. In real-world applications, the obtained intervals could be banned from excavation to ensure that the buried cables are not damaged.

\bibliography{ref}

% Generated by IEEEtran.bst, version: 1.14 (2015/08/26)
\begin{thebibliography}{10}
\providecommand{\url}[1]{#1}
\csname url@samestyle\endcsname
\providecommand{\newblock}{\relax}
\providecommand{\bibinfo}[2]{#2}
\providecommand{\BIBentrySTDinterwordspacing}{\spaceskip=0pt\relax}
\providecommand{\BIBentryALTinterwordstretchfactor}{4}
\providecommand{\BIBentryALTinterwordspacing}{\spaceskip=\fontdimen2\font plus
\BIBentryALTinterwordstretchfactor\fontdimen3\font minus
  \fontdimen4\font\relax}
\providecommand{\BIBforeignlanguage}[2]{{%
\expandafter\ifx\csname l@#1\endcsname\relax
\typeout{** WARNING: IEEEtran.bst: No hyphenation pattern has been}%
\typeout{** loaded for the language `#1'. Using the pattern for}%
\typeout{** the default language instead.}%
\else
\language=\csname l@#1\endcsname
\fi
#2}}
\providecommand{\BIBdecl}{\relax}
\BIBdecl

\bibitem{jaw2013locational}
S.~W. Jaw and M.~Hashim, ``Locational accuracy of underground utility mapping
  using ground penetrating radar,'' \emph{Tunnelling and Underground Space
  Technology}, vol.~35, pp. 20--29, 2013.

\bibitem{yatim2014automated}
N.~Yatim, R.~Shauri, and N.~Buniyamin, ``Automated mapping for underground
  pipelines: An overview,'' in \emph{2014 2nd International Conference on
  Electrical, Electronics and System Engineering (ICEESE)}.\hskip 1em plus
  0.5em minus 0.4em\relax IEEE, 2014, pp. 77--82.

\bibitem{daniels2004ground}
D.~Daniels, \emph{Ground Penetrating Radar}.\hskip 1em plus 0.5em minus
  0.4em\relax The Institution of Engineering and Technology, 2004, vol.~1.

\bibitem{butnor2001use}
J.~Butnor, J.~Doolittle, L.~Kress, S.~Cohen, and K.~Johnsen, ``Use of
  ground-penetrating radar to study tree roots in the southeastern united
  states,'' \emph{Tree physiology}, vol.~21, no.~17, pp. 1269--1278, 2001.

\bibitem{shihab2005radius}
S.~Shihab and W.~Al-Nuaimy, ``Radius estimation for cylindrical objects
  detected by ground penetrating radar,'' \emph{Subsurface Sensing Technologies
  and Applications}, vol.~6, no.~2, pp. 151--166, 2005.

\bibitem{illingworth1988survey}
J.~Illingworth and J.~Kittler, ``A survey of the hough transform,''
  \emph{Computer Vision, Graphics, and Image Processing}, vol.~44, no.~1, pp.
  87--116, 1988.

\bibitem{capineri1998advanced}
L.~Capineri, P.~Grande, and J.~Temple, ``Advanced image-processing technique
  for real-time interpretation of ground-penetrating radar images,''
  \emph{International Journal of Imaging Systems and Technology}, vol.~9,
  no.~1, pp. 51--59, 1998.

\bibitem{bookstein1979fitting}
F.~Bookstein, ``Fitting conic sections to scattered data,'' \emph{Computer
  Graphics and Image Processing}, vol.~9, no.~1, pp. 56--71, 1979.

\bibitem{akima1978method}
H.~Akima, ``A method of bivariate interpolation and smooth surface fitting for
  irregularly distributed data points,'' \emph{ACM Transactions on Mathematical
  Software (TOMS)}, vol.~4, no.~2, pp. 148--159, 1978.

\bibitem{porrill1990fitting}
J.~Porrill, ``Fitting ellipses and predicting confidence envelopes using a bias
  corrected kalman filter,'' \emph{Image and Vision Computing}, vol.~8, no.~1,
  pp. 37--41, 1990.

\bibitem{jaw2011accuracy}
S.~W. Jaw and M.~Hashim, ``Accuracy of data acquisition approaches with ground
  penetrating radar for subsurface utility mapping,'' in \emph{Proceedings of
  IEEE International Conference on RF and Microwave Conference (RFM),}.\hskip
  1em plus 0.5em minus 0.4em\relax IEEE, 2011, pp. 40--44.

\bibitem{al2000automatic}
W.~Al-Nuaimy, Y.~Huang, M.~Nakhkash, M.~Fang, V.~Nguyen, and A.~Eriksen,
  ``Automatic detection of buried utilities and solid objects with gpr using
  neural networks and pattern recognition,'' \emph{Journal of applied
  Geophysics}, vol.~43, no.~2, pp. 157--165, 2000.

\bibitem{caorsi2005electromagnetic}
S.~Caorsi and G.~Cevini, ``An electromagnetic approach based on neural networks
  for the gpr investigation of buried cylinders,'' \emph{IEEE Geoscience and
  Remote Sensing Letters}, vol.~2, no.~1, pp. 3--7, 2005.

\bibitem{youn2002automatic}
H.~Youn and C.~Chen, ``Automatic gpr target detection and clutter reduction
  using neural network,'' in \emph{International Conference on Ground
  Penetrating Radar (GPR 2002)}.\hskip 1em plus 0.5em minus 0.4em\relax
  International Society for Optics and Photonics, 2002, pp. 579--582.

\bibitem{pasolli2009automatic}
E.~Pasolli, F.~Melgani, and M.~Donelli, ``Automatic analysis of gpr images: A
  pattern-recognition approach,'' \emph{IEEE Transactions on Geoscience and
  Remote Sensing}, vol.~47, no.~7, pp. 2206--2217, 2009.

\bibitem{gamba2000neural}
P.~Gamba and S.~Lossani, ``Neural detection of pipe signatures in ground
  penetrating radar images,'' \emph{IEEE Transactions on Geoscience and Remote
  Sensing}, vol.~38, no.~2, pp. 790--797, 2000.

\bibitem{delbo2000fuzzy}
S.~Delbo, P.~Gamba, and D.~Roccato, ``A fuzzy shell clustering approach to
  recognize hyperbolic signatures in subsurface radar images,'' \emph{IEEE
  Transactions on Geoscience and Remote Sensing}, vol.~38, no.~3, pp.
  1447--1451, 2000.

\bibitem{maas2013using}
C.~Maas and J.~Schmalzl, ``Using pattern recognition to automatically localize
  reflection hyperbolas in data from ground penetrating radar,''
  \emph{Computers \& Geosciences}, vol.~58, pp. 116--125, 2013.

\bibitem{borgioli2008detection}
G.~Borgioli, L.~Capineri, P.~Falorni, S.~Matucci, and C.~Windsor, ``The
  detection of buried pipes from time-of-flight radar data,'' \emph{IEEE
  Transactions on Geoscience and Remote Sensing}, vol.~46, no.~8, pp.
  2254--2266, 2008.

\bibitem{chen2010probabilistic}
H.~Chen and A.~Cohn, ``Probabilistic conic mixture model and its applications
  to m=mining spatial ground penetrating radar data,'' in \emph{Workshops of
  SIAM Conference on Data Mining}, 2010.

\bibitem{chen2010robust}
------, ``Probabilistic robust hyperbola mixture model for interpreting ground
  penetrating radar data,'' in \emph{The 2010 International Joint Conference on
  Neural Networks (IJCNN)}.\hskip 1em plus 0.5em minus 0.4em\relax IEEE, 2010,
  pp. 1--8.

\bibitem{windsor2014data}
C.~Windsor, L.~Capineri, and P.~Falorni, ``A data pair-labeled generalized
  hough transform for radar location of buried objects,'' \emph{IEEE Geoscience
  and Remote Sensing Letters}, vol.~11, no.~1, pp. 124--127, 2014.

\bibitem{dou2016real}
Q.~Dou, L.~Wei, D.~Magee, and A.~Cohn, ``Real time hyperbolae recognition and
  fitting in gpr data,'' \emph{IEEE Transactions on Geoscience and Remote
  Sensing}, vol.~55, no.~1, pp. 51--62, 2017.

\bibitem{zhou2018automatic}
X.~Zhou, H.~Chen, and J.~Li, ``An automatic gpr b-scan image interpreting
  model,'' \emph{IEEE Transactions on Geoscience and Remote Sensing}, vol.~56,
  no.~6, pp. 3398--3412, 2018.

\bibitem{zhou2019efficient}
X.~Zhou, H.~Chen, and T.~Hao, ``Efficient detection of buried plastic pipes by
  combining gpr and electric field methods,'' \emph{IEEE Transactions on
  Geoscience and Remote Sensing}, vol.~57, no.~6, pp. 3967--3979, 2019.

\bibitem{dou20163d}
Q.~Dou, L.~Wei, D.~R. Magee, P.~R. Atkins, D.~N. Chapman, G.~Curioni, K.~F.
  Goddard, F.~Hayati, H.~Jenks, N.~Metje \emph{et~al.}, ``3d buried utility
  location using a marching-cross-section algorithm for multi-sensor data
  fusion,'' \emph{Sensors}, vol.~16, no.~11, p. 1827, 2016.

\bibitem{jazwinski2007stochastic}
A.~H. Jazwinski, \emph{Stochastic processes and filtering theory}.\hskip 1em
  plus 0.5em minus 0.4em\relax New York: Academic, 1970.

\bibitem{chen2011buried}
H.~Chen and A.~G. Cohn, ``Buried utility pipeline mapping based on multiple
  spatial data sources: a bayesian data fusion approach,'' in \emph{Proceedings
  of the Twenty-Second International Joint Conference on Artificial
  Intelligence (IJCAI)}, vol.~11, 2011, pp. 2411--2417.

\bibitem{zhou2019probabilistic}
X.~Zhou, H.~Chen, and J.~Li, ``Probabilistic mixture model for mapping the
  underground pipes,'' \emph{ACM Transactions on Knowledge Discovery from Data
  (TKDD)}, vol.~13, no.~5, pp. 1--26, 2019.

\bibitem{ma2008development}
B.~Ma and M.~Najafi, ``Development and applications of trenchless technology in
  china,'' \emph{Tunnelling and Underground Space Technology}, vol.~23, no.~4,
  pp. 476--480, 2008.

\bibitem{jiang2019cable}
G.~Jiang, X.~Zhou, J.~Li, and H.~Chen, ``A cable-mapping algorithm based on
  ground-penetrating radar,'' \emph{IEEE Geoscience and Remote Sensing
  Letters}, vol.~16, no.~10, pp. 1630--1634, 2019.

\bibitem{conyers2002ground}
L.~B. Conyers, ``Ground penetrating radar,'' \emph{Encyclopedia of Imaging
  Science and Technology}, 2002.

\bibitem{williams2006gaussian}
C.~K. Williams and C.~E. Rasmussen, \emph{Gaussian Processes for Machine
  Learning}.\hskip 1em plus 0.5em minus 0.4em\relax Massachusetts Institute of
  Technology Publishing (MIT Press), 2006, vol.~2, no.~3.

\end{thebibliography}
\bibliographystyle{IEEEtran}

\end{spacing}
\end{document}